\pdfoutput=1
\documentclass[5p,twocolumn]{elsarticle}

\usepackage[T1]{fontenc}
\usepackage[utf8]{inputenc}
\usepackage{lmodern}
\usepackage{graphicx}
\graphicspath{{figures/}}
\usepackage{amsmath}
\usepackage{amssymb}
\usepackage{amsthm}
\usepackage{booktabs}
\usepackage{multirow}
\usepackage{xcolor}
\usepackage{adjustbox}
\usepackage{microtype}
\usepackage{lineno}

\usepackage[hidelinks]{hyperref}
\biboptions{sort&compress}

\ifdefined\showedits
  \colorlet{neweditcolor}{red}
\else
  \colorlet{neweditcolor}{black}
\fi
\DeclareRobustCommand{\newedit}[1]{\textcolor{neweditcolor}{#1}}
\journal{Neurocomputing}

\begin{document}

\begin{frontmatter}

\title{Frequency-Aware Model Parameter Explorer: A New Attribution Method for Improving Explainability}

\author[cmpbme]{Ali Yavari\corref{cor1}}
\ead{ali.yavari@meduniwien.ac.at}
\author[indep]{Alireza Mohamadi}
\author[sharif]{Elham Beydaghi}
\author[cir,caim]{Philipp Seeb\"{o}ck}
\author[cmpbme]{Rainer A. Leitgeb}

\cortext[cor1]{Corresponding author.}

\affiliation[cmpbme]{organization={Center for Medical Physics and Biomedical Engineering, Medical University of Vienna},
            addressline={W\"{a}hringer G\"{u}rtel 18--20},
            city={Vienna},
            postcode={1090},
            country={Austria}}
\affiliation[indep]{organization={Independent researcher},
            city={Kermanshah},
            country={Iran}}
\affiliation[sharif]{organization={Department of Computer Engineering, Sharif University of Technology},
            addressline={Azadi Avenue},
            city={Tehran},
            country={Iran}}
\affiliation[cir]{organization={MANO Group, CIR Lab, Department of Biomedical Imaging and Image-Guided Therapy, Medical University of Vienna},
            addressline={W\"{a}hringer G\"{u}rtel 18--20},
            city={Vienna},
            postcode={1090},
            country={Austria}}
\affiliation[caim]{organization={Comprehensive Center for AI in Medicine, Medical University of Vienna},
            addressline={Spitalgasse 23},
            city={Vienna},
            postcode={1090},
            country={Austria}}

\begin{abstract}
State-of-the-art attribution methods {such as AttEXplore} rely on adversarial sample generation that applies an all-pass filter across the frequency spectrum{\newedit{: they perturb all frequencies of the input image alike, from the low frequencies that carry coarse shape to the high frequencies that carry edges and textures. However, models rely on low and high frequencies to different degrees, and such an all-pass perturbation cannot adapt to the model being explained, which limits the faithfulness of the resulting attribution maps.} We propose FAMPE (Frequency-Aware Model Parameter Explorer), an attribution method that \newedit{overcomes this limitation by perturbing} the low- and high-frequency bands of the input independently, \newedit{so that the perturbation can be shifted toward the band that matters for the model}. The boundary between the two bands is set for each image \newedit{automatically} from its spectral energy. On ImageNet with five \newedit{convolutional neural network} and Vision Transformer architectures, FAMPE \newedit{significantly} improves the Insertion Score over \newedit{the strongest baseline, AttEXplore,} on four \newedit{of them, by up to 15.4\%,} at essentially the same computational cost. \newedit{How the perturbed frequency bands are weighted matters: the best and the worst balance between them differ by 19--36\% in Insertion Score, and the best balance depends on the architecture. Choosing the balance for each image with a probe that needs no ground truth raises the Insertion Score significantly further on all five architectures, reaching up to 24.4\% above AttEXplore.} Controlling which frequency band is perturbed is therefore a simple, low-cost way to improve attribution faithfulness.}
\end{abstract}

\begin{keyword}
Explainable artificial intelligence \sep attribution methods \sep feature importance \sep frequency-domain analysis \sep adversarial perturbation \sep deep neural networks
\end{keyword}

\end{frontmatter}

\section{Introduction}
Deep neural networks (DNNs) have become integral to high-stakes decision-making across healthcare \citep{nie2015disease, dhar2023challenges}, autonomous systems \citep{jain2015car}, and finance \citep{chong2017deep, pham2017deep} — yet their opaque "black-box" nature poses fundamental challenges to trust and adoption \citep{ali2023explainable}. In healthcare, clinicians require transparent AI recommendations to ensure reliability, yet traditional models often lack clarity \citep{murad2024unraveling, ibrahimov2024explainable}. This opacity complicates bias detection and regulatory compliance, especially in sensitive areas like finance and autonomous systems, where fairness and accountability are critical \citep{maier2022relationship}. Furthermore, the lack of interpretability hinders ethical governance, making it difficult to align AI behaviors with societal values \citep{murad2024unraveling}. These issues highlight the urgent need for methods that elucidate DNN decision-making processes to foster trust and ensure responsible deployment.

Initial XAI methods such as LIME \citep{ribeiro2016whyitrustyou} provide local surrogate explanations for interpreting DNNs but suffer from oversimplification. Gradient-based techniques, including Saliency Maps (SM), Grad-CAM \citep{Selvaraju_2017_ICCV}, and Score-CAM \citep{wang2020scorecam}, improve interpretability but lack fine-grained precision. Integrated Gradients (IG) \citep{sundararajan2017axiomatic} advances the field with axiomatic guarantees, and more recent adversarial-sample-based methods such as Adversarial Gradient Integration (AGI) \citep{pan2021explaining}, Boundary-based IG (BIG) \citep{wang2021robust}, MFABA \citep{zhu2024mfaba}, and AttEXplore \citep{zhu2024attexplore} achieve higher accuracy by exploring decision boundaries, albeit with added computational cost.

\begin{figure*}[!t]
  \centering
  \includegraphics[width=0.9\linewidth]{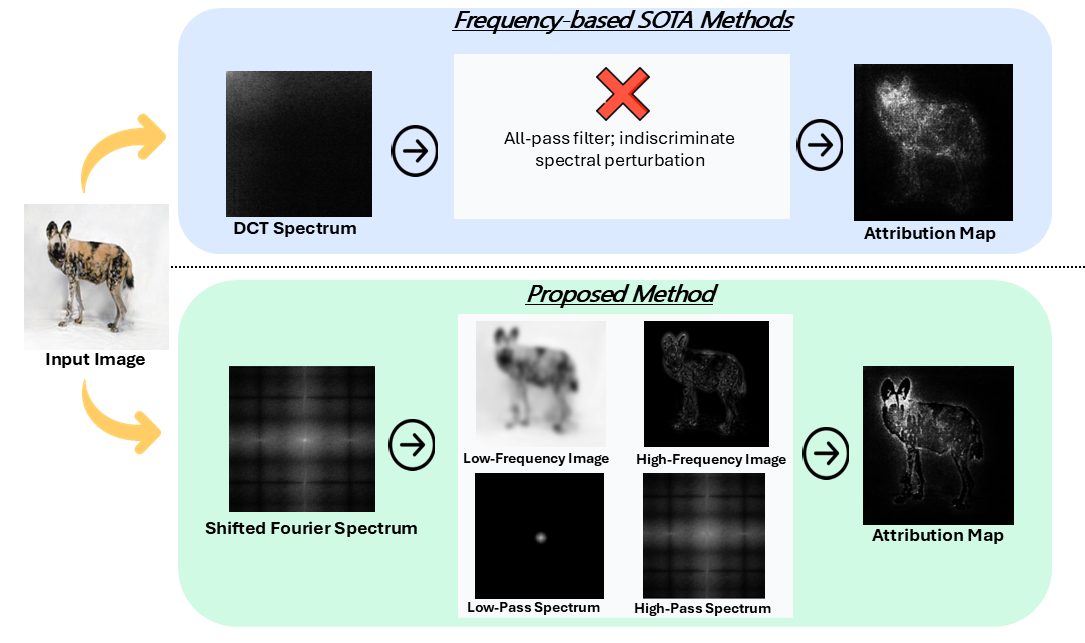}
  \caption{Comparison between frequency-based SOTA attribution methods and our proposed approach. \textit{Top:} Existing methods such as AttEXplore apply a single all-pass multiplicative perturbation to the whole spectrum, so every frequency band is perturbed alike and no band can be weighted separately. \textit{Bottom:} FAMPE splits the spectrum at an energy-defined cutoff and perturbs the low-frequency band (capturing the object's overall shape, e.g., the dog's silhouette) and the high-frequency band (capturing edges and fine details) with two independently sampled fields weighted by $\alpha$ and $1-\alpha$. We use a centered Fourier spectrum because the cutoff is defined as a radius from the DC component; the same construction could be built on a DCT. In this example the resulting map is more closely aligned with the object; the population-level comparison is given in \newedit{Section~\ref{sec:results}}.}
  \label{fig1}
\end{figure*}

Among current state-of-the-art attribution methods, AttEXplore \citep{zhu2024attexplore} improves attribution by combining model parameter exploration (MPE) with frequency-based feature modifications, producing more faithful attribution maps than gradient-based methods such as IG \citep{sundararajan2017axiomatic} and GIG \citep{kapishnikov2021guided}. However, as illustrated in the top row of Fig.~\ref{fig1}, AttEXplore {perturbs the entire spectrum at once with a single all-pass noise field, so low and high frequencies cannot be perturbed separately.} This matters because the two frequency regimes carry fundamentally different visual information: low frequencies describe an object's overall shape and global layout, while high frequencies encode the edges, textures, and fine details that often drive a model's prediction~\citep{wang2020highfrequency, yin2019fourier}. A method that cannot probe these regimes independently is therefore blind to an entire class of features that may underlie the model's decision. Our proposed method addresses this gap {by splitting the spectrum into a low- and a high-frequency band and perturbing each band independently. In the example of Fig.~\ref{fig1} (bottom row), this yields a map that is more closely aligned with the object.} To address this gap formally, we pose the following research questions: {\textbf{(i)} Does separating high- and low-frequency components make attribution maps more faithful, as measured by Insertion and Deletion~\citep{petsiuk2018rise}? \textbf{(ii)} Do these gains carry over to localization against ground-truth bounding boxes? \textbf{(iii)} What is the role of each frequency band?}

To address these questions, we propose FAMPE (Frequency-Aware Model Parameter Explorer), an attribution method that disentangles the high- and low-frequency components of the input spectrum during transferable adversarial perturbation, generating adversarial samples that expose which spectral band most strongly drives model predictions. While AttEXplore \citep{zhu2024attexplore} demonstrates that transferable adversarial attacks can drive attribution by exploring decision boundaries across models, it does so using an all-pass filter that treats the full frequency spectrum uniformly. FAMPE introduces \textit{frequency-aware} transferable perturbations — separately modulating high- and low-frequency components — thereby allowing the attribution process to selectively probe which spectral band carries the most decision-relevant information. This is enabled by a tunable parameter $\alpha$ that continuously interpolates between a purely high-frequency and a purely low-frequency perturbation regime, giving practitioners explicit control over which spectral band is explored and allowing a systematic empirical investigation of how each frequency band of the perturbation contributes to attribution quality. {We analyze how $\alpha$ shapes the resulting attribution maps across image complexities and architectures, using Insertion and Deletion scores, localization against ground-truth bounding boxes, and qualitative heatmaps. The results show that frequency-aware exploration improves attribution faithfulness on most architectures, while its effect on localization depends on the model.}

{The main contributions of this paper are as follows.
\begin{itemize}
    \item FAMPE introduces a frequency-aware perturbation for adversarial attribution: two independently sampled fields on energy-defined bands, weighted by $\alpha$ and $1-\alpha$, with the band boundary set per image by an energy-driven spectral cutoff. To our knowledge, FAMPE is the first adversarial-sample-based attribution method to control which frequency band the perturbation targets. Doing so shows that the band strongly affects the attribution and that the best band depends on the model being explained; unlike the low-pass gradient filter of FORGrad~\citep{muzellec2024saliency}, the best balance favors the high-frequency band on two of five architectures.
    \item We introduce a cheap, label-free $\alpha$ selector: a truncated Insertion probe that chooses $\alpha$ per image at test time, using only the model and the input image.
    \item Extensive experiments on ImageNet with five CNN and Vision Transformer architectures show that FAMPE improves the Insertion Score over AttEXplore by up to 15.4\% on four architectures and ties on the fifth, at essentially the same cost. The selector raises it further on all five, reaching 24.4\% above AttEXplore on MaxViT-T. Localization against ground-truth bounding boxes \newedit{improves most where the all-pass baseline localizes worst, by up to 8.1 points in the pointing game on MaxViT-T. On the other architectures, higher Insertion does not consistently translate into better localization, since the two measure different things: Insertion measures agreement with the model, localization agreement with the annotated object. Attribution methods tuned on Insertion should therefore also be validated with an independent measure.} Ablations show that the best $\alpha$ is nearly independent of the image's spectral cutoff.
\end{itemize}}
\section{Related Work}
We organize strategies for interpreting DNNs into two groups: traditional methods (local approximation and gradient-based attribution) and advanced methods (adversarial-sample-based attribution). 

\subsection{Local approximation methods}
Local approximation techniques aim to interpret the behavior of complex models in the vicinity of specific inputs by building simpler, more interpretable surrogate models. A prominent example is LIME \citep{ribeiro2016whyitrustyou}, which achieves local interpretability by fitting interpretable surrogate models around a given sample, but relies on assumptions of local linearity and feature independence that often fail for highly nonlinear DNNs, and incurs substantial computational cost since a new surrogate must be fitted for every input. Subsequent approaches include Layer-wise Relevance Propagation (LRP) \citep{Binder2016LayerWiseRP} and DeepLIFT \citep{Shrikumar_deeplift}. LRP assigns relevance scores by propagating the prediction backward through layers, but its results are highly architecture-dependent. DeepLIFT evaluates feature importance by comparing input values with predefined reference points; however, its sensitivity to the choice of reference points can introduce inconsistencies, and it does not satisfy the Implementation Invariance axiom later formalized by Integrated Gradients (IG) \citep{sundararajan2017axiomatic}, which may result in biased explanations.

\subsection{Gradient-based attribution methods}
Since neural networks are trained using gradients, gradient-based approaches naturally emerged for interpreting model predictions. The earliest such method, Saliency Maps (SM) \citep{simonyan2014deepinsideconvolutionalnetworks}, identifies influential features by computing the gradient of the output with respect to the input, but suffers from gradient saturation and fails to satisfy the Sensitivity axiom later formalized in IG \citep{sundararajan2017axiomatic}, which can result in zero attribution despite changes in model output. To overcome these limitations, IG accumulates gradients along a path from a baseline to the input, satisfying both the Sensitivity and Implementation Invariance axioms that serve as key guarantees for attribution methods. However, IG is computationally expensive due to its multiple forward and backward passes, and is further constrained by the choice of baseline, which often introduces irrelevant noise into the attributions. {This baseline dependence is not a minor implementation detail: Sturmfels et al.~\citep{sturmfels2020baselines} show that different, individually reasonable baselines (black, white, uniform noise, blurred, or a maximum-entropy distribution) can yield IG attributions that emphasize entirely different regions of the same image, so the resulting explanation reflects the analyst's choice of reference point as much as it reflects the model. Adversarial-sample-based methods such as AGI and AttEXplore sidestep the need for a fixed hand-picked baseline entirely, deriving perturbation directions from the model's own decision boundary instead.} Variants such as Fast IG \citep{hesse2021fast} and Guided IG \citep{kapishnikov2021guided} attempt to address these issues — the former by improving numerical integration efficiency at the cost of approximation errors, and the latter by selectively propagating gradients, though it tends to overemphasize features tied to specific categories while neglecting indirect ones. A separate line of work, including Grad-CAM \citep{Selvaraju_2017_ICCV} and Score-CAM \citep{wang2020scorecam}, leverages gradient information from intermediate layers to localize important regions; however, their explanations remain coarse and lack the fine-grained resolution required of true attribution methods. Closest to the spectral view taken here, FORGrad~\citep{muzellec2024saliency} traces the noise in white-box attribution maps to high-frequency artifacts in the gradient, caused mainly by max-pooling and striding, and removes them with a low-pass filter applied to each gradient before it is turned into an attribution map. Its cutoff is fixed once per architecture and attribution method rather than per image, and it can only suppress high frequencies. FAMPE instead acts on the input variants that the exploration evaluates, can emphasize either band, sets the band boundary per image from its spectral energy, and can also choose the band weighting per image (Sec.~\ref{sec:selection}).

\subsection{Adversarial-sample-based attribution methods}
Rather than relying on a manually predefined reference point, adversarial-sample-based attribution methods generate adversarial instances and probe decision boundaries, letting the model itself reveal which features are decisive by perturbing inputs until the prediction changes. AGI \citep{pan2021explaining} exemplifies this class by applying targeted adversarial perturbations in the input space to investigate boundary behavior and refine attributions through nonlinear path-integrated gradients. Despite its innovation, AGI's reliability is constrained by the stability and quality of adversarial examples, as it may traverse numerous decision boundaries that are not always relevant to the explanation. BIG \citep{wang2021robust} automates the search for a suitable reference point by locating it on the model's decision boundary, removing the need for manual specification while still producing more precise feature attributions. Nonetheless, its use of a linear integration path may restrict its capacity to fully capture the nonlinear and complex behavior of model decisions. More recently, MFABA (More Faithful and Accelerated Boundary-based Attribution) \citep{zhu2024mfaba} improves both accuracy and efficiency through second-order Taylor expansion and boundary exploration, making it well-suited for complex models, though computational costs increase due to higher-order derivative calculations. AttEXplore \citep{zhu2024attexplore} leverages model parameter exploration and frequency-based feature alterations to generate precise attribution maps. Although AttEXplore provides promising results, it relies on an all-pass filter to generate frequency-based adversarial examples, which does not separate high and low frequency components. {As a result, the exploration cannot emphasize the band that is most informative for a given model, whether it carries fine detail (high frequencies) or coarse shape (low frequencies).} {The all-pass spectral perturbation used by AttEXplore originates in the spectrum simulation attack of Long et al.~\citep{long2022ssa}, where it was designed to improve adversarial transferability. AttEXplore transferred it to attribution, and the subsequent extension of Zhu et al.~\citep{zhu2026improving} varies the gradient-acquisition operator while leaving the all-pass perturbation intact. FAMPE is, to our knowledge, the first adversarial-sample-based attribution method to control which frequency band the perturbation targets.}
\section{Method}

\begin{figure*}[!t]
  \centering
  \includegraphics[width=\linewidth]{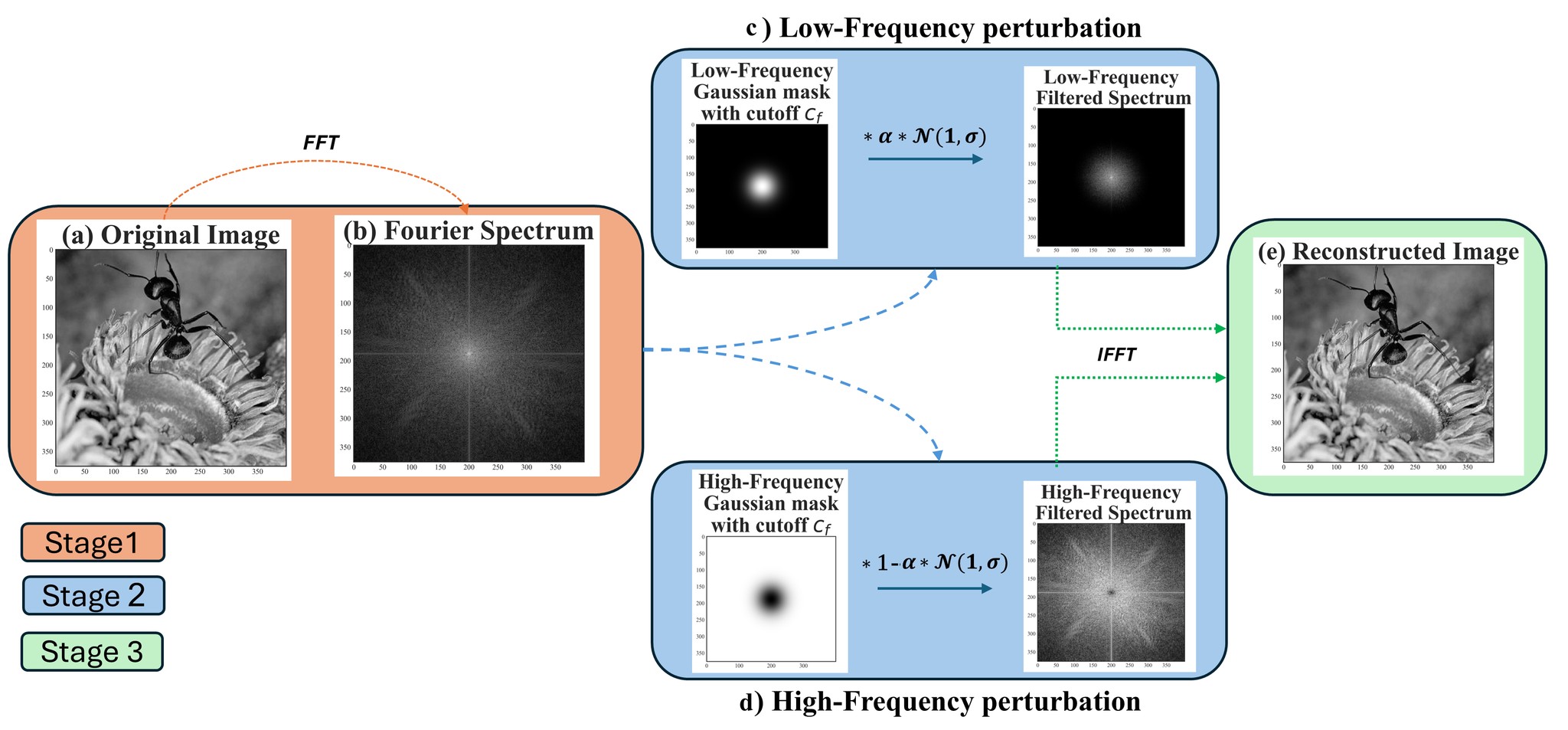}
  \caption{Overview of FAMPE's frequency-aware adversarial sample generation. Stage 1: input image (a) is transformed via FFT to obtain the centered Fourier spectrum (b). Stage 2: the spectrum is independently modulated by a low-frequency Gaussian mask scaled by $\alpha$ (c, top) and a high-frequency Gaussian mask scaled by $1-\alpha$ (d, bottom), each combined with multiplicative Gaussian noise $\mathcal{N}(1,\sigma)$. Stage 3: modulated spectra are summed and mapped back via IFFT, yielding the reconstructed perturbed image (e). The parameter $\alpha$ controls the relative perturbation per spectral band.}
  \label{fig2}
\end{figure*}

FAMPE is built on the principle that DNN decisions are encoded across distinct frequency bands of the input — low frequencies capturing the global spatial layout and overall form of objects, and high frequencies encoding edges, textures, and fine structural details. Building on \newedit{the} insight that input perturbations can be used as a proxy for model parameter exploration to drive attribution\newedit{~\citep{long2022ssa}}, FAMPE conducts this exploration in a \textit{frequency-aware} manner. At the core of the method is a frequency-aware adversarial sample generation procedure, illustrated in Fig.~\ref{fig2} and formalized in \newedit{Section~\ref{sec:faperturb}}, which proceeds in three stages: (1) \textit{spectral decomposition} of the input via the FFT, (2) \textit{$\alpha$-weighted frequency-aware perturbation}, in which the low- and high-frequency components of the spectrum are independently modulated by multiplicative Gaussian noise scaled by $\alpha$ and $1-\alpha$ respectively, and (3) \textit{spatial reconstruction} via the inverse FFT to produce the perturbed sample. By varying $\alpha$, this procedure continuously interpolates between purely low-frequency and purely high-frequency perturbation regimes, allowing FAMPE to selectively probe which spectral band carries the most decision-relevant information for a given input and model. The resulting perturbed samples are then used in a nonlinear gradient-integration scheme \newedit{(also Section~\ref{sec:faperturb})} that accumulates per-feature sensitivity across \newedit{$N$} such samples to yield the final attribution map. {Sec.~\ref{sec:selection} adds an optional selector that chooses $\alpha$ per image.}

\subsection{Axiomatic Properties}
FAMPE inherits from AttEXplore~\citep{zhu2024attexplore} its relation to the axioms of Sundararajan \emph{et al.}~\citep{sundararajan2017axiomatic}. It satisfies \textit{Implementation Invariance} when functionally equivalent networks are run with the same random draws and are differentiable at the evaluated points, because it uses only the networks' outputs and input gradients. Under the stated assumptions, namely a first-order expansion along the exploration path and the gradient averaged over the perturbed variants approximated by the gradient at the iterate, an approximate completeness relation follows: the attributions sum to the change in loss along the path. This relation does not establish \textit{Sensitivity}, which concerns an input and a baseline that differ in a single feature; FAMPE has no such baseline, since the endpoint of its exploration generally differs from the input in many features. The derivation is given in \ref{sec:appendix_axioms}.

\subsection{Frequency-Aware Adversarial Perturbation}
\label{sec:faperturb}

We now describe the core of FAMPE: the construction of frequency-aware adversarial samples that drive the attribution process. The intuition is that perturbing high- and low-frequency components separately, rather than uniformly across the spectrum, reveals which spectral band most strongly influences the model's decision, and the parameter $\alpha$ provides explicit control over this trade-off.
The attribution itself is accumulated along a nonlinear path, as in AGI and AttEXplore~\citep{pan2021explaining,zhu2024attexplore}; FAMPE inherits this integration form unchanged, Eq.~\eqref{eq:path_eq},
\begin{equation}
A = \int \Delta x^t \odot g(x^t) \, dt
\label{eq:path_eq}
\end{equation}
in which $\Delta x^t$ is the displacement of the iterate toward the decision boundary at step $t$, $g(x^t)$ is the gradient signal collected at that step, and $\odot$ is the elementwise product. The integral therefore credits each input feature with its cumulative contribution to reaching the boundary. What distinguishes FAMPE is not this form but the samples on which $\Delta x^t$ and $g(x^t)$ are evaluated, which we now define.

At every iteration $t$ we draw $N$ frequency-aware variants of the current iterate. The $i$-th variant, $x^{t}_{\omega_i}$, is obtained from Eq.~\eqref{eq:main_eq}, with every random quantity sampled independently for each $i$:

\begin{multline}
x^{t}_{\omega_i} = \mathrm{IFFT}\!\Bigl(\mathrm{FFT}\!\bigl(x^t + \mathcal{N}(0,1)\cdot\epsilon/255\bigr) \\
{}\times \bigl(\alpha \cdot \mathrm{LfM}(c_f) \cdot \mathcal{N}(1,\sigma) + (1-\alpha) \cdot \mathrm{HfM}(c_f) \cdot \mathcal{N}(1,\sigma)\bigr)\Bigr).
\label{eq:main_eq}
\end{multline}

The construction has three stages. First, spatial Gaussian noise $\mathcal{N}(0,1)\cdot \epsilon/255$ is added to $x^t$, with the perturbation rate $\epsilon$ setting its magnitude. Second, the noisy image is mapped to the frequency domain by the fast Fourier transform (FFT)~\citep{cooley1965algorithm} and shifted so that the low frequencies sit at the center of the spectrum, where it is split by two complementary masks: the low-frequency mask $\mathrm{LfM}(c_f)$, a Gaussian mask passing frequencies below the cutoff $c_f$, is modulated by a multiplicative noise field $\mathcal{N}(1, \sigma)$ and weighted by $\alpha$, while the high-frequency mask $\mathrm{HfM}(c_f)$, its complement, is modulated by a second, independently drawn field $\mathcal{N}(1, \sigma)$ and weighted by $1-\alpha$. The parameter $\sigma$ sets the strength of this multiplicative noise, and $c_f$ is computed per image as described in Section~\ref{sec:cutoff}. Third, the inverse FFT (IFFT) returns the result to the spatial domain. The parameter $\alpha$ sets the gains of the two perturbed bands, scaling signal and noise alike; $\alpha=0$ retains only the high-pass component and $\alpha=1$ only the low-pass component. Fig.~\ref{fig2} illustrates the three stages. From the $N$ variants, the update direction $\Delta x^t$ and the aggregated gradient $g(x^t)$ follow as Eqs.~\eqref{eq:minor_eq_one} and~\eqref{eq:minor_eq_two},

\begin{equation}
\Delta x^t = \eta \cdot \mathrm{sign}\!\left(\frac{1}{N}\sum_{i=1}^{N} \frac{\partial L\!\left(x^{t}_{\omega_i},y\right)}{\partial x^{t}_{\omega_i}}\right),
\label{eq:minor_eq_one}
\end{equation}
\begin{equation}
g(x^t) = \frac{1}{N}\sum_{i=1}^{N} \frac{\partial L\!\left(x^{t}_{\omega_i},y\right)}{\partial x^{t}_{\omega_i}},
\label{eq:minor_eq_two}
\end{equation}

where $L(\cdot,y)$ is the classification loss for the original label $y$ and $\eta$ is the {learning rate}. Both quantities are built from the same mean gradient over the $N$ variants: taking its sign gives the step direction $\Delta x^t$, which advances the iterate toward the decision boundary by a fixed amount per coordinate, while the mean itself, $g(x^t)$, serves as the per-feature importance signal at that step. Averaging over $N$ independently perturbed variants makes both estimates robust to any single draw of the noise. Their elementwise product, integrated along the trajectory by Eq.~\eqref{eq:path_eq}, yields the attribution map.

\subsection{Energy-Based Cutoff Determination}
\label{sec:cutoff}

The cutoff $c_f$ defines the boundary between low- and high-frequency regions in Eq.~\eqref{eq:main_eq}. {The meaningful spectral split varies substantially across images: uniform scenes concentrate their energy in a narrow low-frequency core, while highly textured scenes require a much larger cutoff to capture their structural content. We therefore determine $c_f$ automatically from the spectral energy distribution of each image.} The image is transformed into the frequency domain via FFT, yielding the complex spectrum $F(u,v)$. The cutoff $c_f$ is then defined as the smallest radius $r$ for which the cumulative spectral energy within radius $r$ reaches a specified fraction $\tau$ of the total energy (excluding the DC component at $r=0$), as formalized in Eq.~\eqref{eq:main_cutoff} 

\begin{multline}
c_f = \min\biggl\{r \;\bigg|\; \sum_{0 < \sqrt{u^2+v^2} \leq r} |F(u,v)|^2 \\
\geq\; \tau \sum_{\sqrt{u^2+v^2} > 0} |F(u,v)|^2 \biggr\}.
\label{eq:main_cutoff}
\end{multline}

This guarantees that the low-frequency region enclosed by $c_f$ captures the majority of the image's energy, adapting the spectral split to each input.

\subsection{Per-Image \texorpdfstring{$\alpha$}{alpha} Selection}
\label{sec:selection}
{\nopagebreak By default, FAMPE uses a single fixed $\alpha^*$ per architecture, the best value in the $\alpha$ sweep. The best $\alpha$, however, also varies from image to image and is nearly independent of the image's spectral cutoff (Sec.~\ref{sec:ablation}). We therefore propose an optional selector that chooses $\alpha$ per image by measuring the model's response to each candidate directly. For each \newedit{candidate value of $\alpha$ on a grid over $[0,1]$}, a truncated FAMPE run with only a few attack steps and approximate features produces a coarse attribution map. Each coarse map is scored with a coarse version of the Insertion Score~\citep{petsiuk2018rise}, which measures how quickly the model's confidence recovers as pixels are inserted in order of decreasing attribution; the coarse version evaluates only a few insertion fractions. The candidate with the highest score is selected, and the final attribution is computed once, at full cost, with that $\alpha$. Selection uses only the model and the input image: it has no access to ground-truth class labels or bounding boxes. It does, however, use a truncated form of the Insertion metric we report, so an Insertion gain under selection is not by itself independent evidence of a better explanation. We therefore use localization against ground-truth bounding boxes as an independent check (Sec.~\ref{sec:results}).}
\section{Experiments and Results}
\subsection{Experimental Setup}
\label{sec:setup}

\noindent \textbf{Dataset and Backbones.} \hspace{0.5em} Our approach is tested on ImageNet \citep{deng2009imagenet}, comprising more than 14 million samples. To keep the experiments tractable while still representative, we randomly select 1000 samples following guidelines from prior works \citep{zhu2024attexplore,pan2021explaining}, {matching the evaluation scale of the two closest baselines, AGI and AttEXplore}. Moreover, we evaluate our proposed attribution method on a selection of widely used CNN architectures, including Inception-v3 \citep{szegedy2016rethinking}, ResNet-50 \citep{he2016deep}, and VGG-16 \citep{simonyan2014very}, as well as {two Transformer-based models, MaxViT-T \citep{tu2022maxvit} and ViT-B/16 \citep{dosovitskiy2021image}}, to assess its performance across different model types.

\noindent \textbf{Baselines.} \hspace{0.5em} As baselines, we compare our method with two state-of-the-art attribution methods, AGI \citep{pan2021explaining} and AttEXplore \citep{zhu2024attexplore}, alongside three classical techniques: IG \citep{sundararajan2017axiomatic}, DeepLIFT \citep{Shrikumar_deeplift}, and GIG \citep{kapishnikov2021guided}. {We compare against the original AttEXplore rather than its extension \citep{zhu2026improving}. The extension varies the gradient-acquisition operator and leaves the all-pass perturbation intact, so it is orthogonal to the band separation studied here. All baselines are evaluated on the same $n=1000$ images per architecture as FAMPE, so all Insertion and Deletion comparisons in Table~\ref{tab:overall} are computed on identical, paired samples.}

\noindent \textbf{Metrics.} \hspace{0.5em} To measure performance, we adopt the Insertion and Deletion Scores \citep{petsiuk2018rise}. The Insertion Score measures the recovery of model predictions when key features are reintroduced, with higher values often associated with greater interpretability. The Deletion Score reflects the model's sensitivity to the removal of important features, where lower values can be indicative of stronger attribution performance. Following established practice in the attribution literature \citep{petsiuk2018rise,zhu2024attexplore}, we treat Insertion Score as the primary evaluation metric, since the Deletion Score can yield misleading signals due to the adversarial sensitivity of neural networks --- specifically, removing pixels can produce out-of-distribution inputs whose model responses do not faithfully reflect feature importance. The Deletion Score is therefore reported as a complementary, secondary indicator.

\noindent \textbf{Parameters.} \hspace{0.5em} The framework is implemented in Python 3.11 with TensorFlow 2.18, utilizing an Intel Core i9 CPU, and all experiments are executed on an NVIDIA GeForce RTX 4090 GPU (24 GB). For a fair comparison with AttEXplore, both methods are evaluated under identical settings: learning rate {$\eta = 0.05$}, number of approximate features $N = 20$, {Gaussian noise standard deviation $\sigma = 16/255$}, perturbation rate $\epsilon = 48$, and a total of 10 attack iterations. {Our AttEXplore implementation follows the equations of the published paper \citep{zhu2024attexplore}. To check that FAMPE differs from it only in the band split, we also run FAMPE with the split disabled (all-pass control).} In addition, we use a Gaussian mask and calculate the cutoff point for every image based on the energy function by setting $\tau = 0.9$ to represent 90\% of the total energy. {We report three FAMPE configurations: a fixed $\alpha^*$, one value per architecture shared by all images, which is the primary comparison; a \emph{selected} $\alpha$ chosen per image at test time by the probe of Sec.~\ref{sec:selection}; and a per-sample oracle upper bound, for which the maximum Insertion and minimum Deletion scores are extracted per sample\newedit{, each over the full sweep and independently of the other,} and averaged. The probe \newedit{evaluates the 11 candidates $\alpha \in \{0, 0.1, \ldots, 1.0\}$, the same grid as the $\alpha$ ablation; it} runs 2 attack steps with 5 approximate features per candidate (10 gradient evaluations instead of 200) and scores each coarse map with Insertion on an 11-point grid of insertion fractions instead of the 101-point grid used for reporting. Where we compare FAMPE against AttEXplore, or selected $\alpha$ against fixed $\alpha^*$, we use a paired bootstrap over images (10{,}000 resamples, 95\% CI); ``significant'' means the interval excludes zero. Insertion and Deletion scores are reported as means, and significance is stated in the text. $\alpha^*$ is the value with the highest mean Insertion Score in the $\alpha$ ablation \newedit{(Section~\ref{sec:ablation})}. The mean Insertion gains are not confined to this optimum: on each of the four architectures where FAMPE improves on AttEXplore, the neighboring values $\alpha^* \pm 0.1$ also exceed AttEXplore's mean Insertion Score \newedit{(Table~\ref{tab:overall} and Section~\ref{sec:ablation})}.}

{
\noindent \textbf{Localization protocol.} \hspace{0.5em} We also score attribution maps against ImageNet ground-truth bounding boxes \citep{deng2009imagenet} on the same $n=1000$ images per architecture. Unlike Insertion, which is used to select $\alpha$ and can disagree with expert visual assessment \citep{Yavari2026}, this check does not depend on the model's predictions. We report two localization measures. The \emph{pointing game} \citep{zhang2018top} scores an attribution map $A$ as a hit if $\arg\max |A|$, the single pixel of peak absolute attribution, falls inside \emph{any} of the image's ground-truth boxes, and a miss otherwise; the reported score is the hit rate averaged over images. The \emph{energy ratio} \citep{wang2020scorecam} instead reports, for each image, the fraction of the map's total attribution mass that falls inside the ground-truth boxes, $\sum_{(i,j) \in \mathrm{boxes}} |A_{ij}| \big/ \sum_{i,j} |A_{ij}|$, giving a continuous score that is not driven by a single argmax pixel. Ground-truth boxes are given in original-image pixel coordinates and are rescaled independently along each axis to match the network's input resolution, since resizing does not preserve aspect ratio. \newedit{As a chance-level reference, we also score a random attribution map (Table~\ref{tab:loc}). Its energy ratio equals the mean fraction of the image covered by the boxes on every architecture, as expected if box rescaling and normalization are correct.} We compare methods with the same paired bootstrap used elsewhere in this section (10{,}000 resamples, seed 42).
}

{
\noindent \textbf{Computational cost.} \hspace{0.5em} We measure per-sample cost in two ways: the number of forward and backward passes implied by each method's protocol, which is the same on every architecture, and the wall-clock time per image on VGG-16 on an otherwise idle RTX 4090.
}

\subsection{Results}
\label{sec:results}

\nopagebreak
\noindent \textbf{Quantitative comparison.} \hspace{0.5em} FAMPE with a single $\alpha^*$ per architecture outperforms AttEXplore on Insertion on four of five architectures and matches it on the fifth (Table~\ref{tab:overall}). The gains are largest on MaxViT-T ($+15.4\%$) and ViT-B/16 ($+5.5\%$), followed by VGG-16 ($+4.1\%$) and ResNet-50 ($+3.7\%$), each significant by paired bootstrap; on Inception-v3 the difference is not significant ($0.5314$ vs.\ $0.5316$). Averaged over the five architectures, Insertion rises from $0.4031$ to $0.4249$. This gain is associated with the complete frequency-weighted procedure: with the split disabled, FAMPE reproduces AttEXplore to within $\pm0.0026$ on every architecture. Choosing $\alpha$ per image improves Insertion further on all five architectures, by $0.023$ to $0.048$ over the fixed $\alpha^*$ (each significant), and raises the mean to $0.4600$. \newedit{For reference, Table~\ref{tab:overall} also reports a per-image oracle, which picks for each image the $\alpha$ with the highest Insertion Score in the full sweep. The oracle uses the evaluation score itself and therefore cannot be deployed, but it shows the best result that any per-image choice among the 11 candidates could achieve. The selector attains $37$--$83\%$ of the oracle's improvement over the fixed $\alpha^*$, depending on the architecture; on VGG-16, for example, the oracle improves Insertion by $0.052$ over the fixed $\alpha^*$ and the selector by $0.043$.} On the secondary metric, FAMPE lowers the Deletion Score relative to AttEXplore on four of five architectures; on ViT-B/16 the difference is not significant. All other baselines score below both AttEXplore and FAMPE on Insertion on every architecture, although AGI achieves the lowest Deletion Score of all methods on ResNet-50 and VGG-16.

\noindent \textbf{Localization.} \hspace{0.5em} \newedit{Averaged over the five architectures, FAMPE localizes better than AttEXplore on both metrics and in both configurations (Table~\ref{tab:loc}, Mean column). The pointing game rises from $0.7536$ to $0.7636$ with the fixed $\alpha^*$ and to $0.7744$ with the selected $\alpha$; the energy ratio rises from $0.6821$ to $0.6900$ and $0.6974$. This average gain comes mainly from MaxViT-T, and per architecture the Insertion gains carry over only in part (Table~\ref{tab:loc_diff}).} At the fixed $\alpha^*$, FAMPE localizes significantly better than AttEXplore on MaxViT-T (pointing $+4.8$ points, energy ratio $+0.044$) and ResNet-50 (energy ratio), and significantly worse on Inception-v3 (pointing) and VGG-16 (energy ratio); the other five of the ten comparisons show no significant difference. Per-image selection improves localization over the fixed $\alpha^*$ in three of ten comparisons (MaxViT-T on both metrics, VGG-16 energy ratio) and worsens it in one (Inception-v3 energy ratio), so on Inception-v3 the selected configuration is significantly behind AttEXplore on both metrics.

\begin{table*}[!t]
\centering

\resizebox{\textwidth}{!}{%
\begin{tabular}{lcccccccccccc}
\toprule
\multirow{2}{*}{\textbf{Method}}
  & \multicolumn{2}{c}{\textbf{Inception-v3}}
  & \multicolumn{2}{c}{\textbf{ResNet-50}}
  & \multicolumn{2}{c}{\textbf{VGG-16}}
  & \multicolumn{2}{c}{\textbf{MaxViT-T}}
  & \multicolumn{2}{c}{\textbf{ViT-B/16}}
  & \multicolumn{2}{c}{\textbf{Mean}} \\
\cmidrule(lr){2-3} \cmidrule(lr){4-5} \cmidrule(lr){6-7} \cmidrule(lr){8-9} \cmidrule(lr){10-11} \cmidrule(lr){12-13}
 & \textbf{Ins.} & \textbf{Del.}
 & \textbf{Ins.} & \textbf{Del.}
 & \textbf{Ins.} & \textbf{Del.}
 & \textbf{Ins.} & \textbf{Del.}
 & \textbf{Ins.} & \textbf{Del.}
 & \textbf{Ins.} & \textbf{Del.} \\
\midrule
IG & 0.1990 & 0.1650 & 0.1853 & 0.1402 & 0.1360 & 0.1123 & 0.3342 & 0.2893 & 0.2547 & 0.2197 & 0.2219 & 0.1853 \\
DeepLIFT & 0.1975 & 0.1769 & 0.1820 & 0.1637 & 0.1415 & 0.1170 & 0.3360 & 0.2890 & 0.2560 & 0.2200 & 0.2226 & 0.1933 \\
GIG & 0.2076 & 0.1483 & 0.1912 & 0.1202 & 0.1409 & 0.0916 & 0.3506 & 0.2606 & 0.2544 & 0.1967 & 0.2289 & 0.1635 \\
AGI & 0.3839 & 0.1169 & 0.3100 & \textbf{0.0975} & 0.2487 & \textbf{0.0774} & 0.2928 & 0.3011 & 0.4350 & 0.1861 & 0.3341 & 0.1558 \\
AttEXplore & 0.5316 & 0.1027 & 0.3448 & 0.1068 & 0.2795 & 0.0899 & 0.3829 & 0.2488 & 0.4766 & 0.1572 & 0.4031 & 0.1411 \\
FAMPE (all-pass, control) & 0.5312 & 0.1029 & 0.3464 & 0.1061 & 0.2805 & 0.0902 & 0.3824 & 0.2501 & 0.4740 & 0.1564 & 0.4029 & 0.1411 \\
\textbf{FAMPE ($\alpha^*$ fixed)} & 0.5314 & \textbf{0.0990} & 0.3575 & 0.1020 & 0.2909 & 0.0837 & 0.4419 & 0.1974 & 0.5028 & 0.1613 & 0.4249 & 0.1287 \\
\textbf{FAMPE (selected $\alpha$)} & \textbf{0.5541} & 0.1004 & \textbf{0.4053} & 0.1021 & \textbf{0.3339} & 0.0853 & \textbf{0.4764} & \textbf{0.1842} & \textbf{0.5303} & \textbf{0.1434} & \textbf{0.4600} & \textbf{0.1231} \\
\textit{FAMPE (oracle $\alpha$)} & \textit{0.5860} & \textit{\newedit{0.0639}} & \textit{0.4230} & \textit{\newedit{0.0696}} & \textit{0.3430} & \textit{\newedit{0.0593}} & \textit{0.4983} & \textit{\newedit{0.1455}} & \textit{0.5772} & \textit{\newedit{0.0887}} & \textit{0.4855} & \textit{\newedit{0.0854}} \\
\bottomrule
\end{tabular}%
}
\caption{Performance of our method, FAMPE, compared to five competing approaches across five deep learning architectures (n=1000 images per cell). Evaluation uses Insertion Score ($\uparrow$ better) and Deletion Score ($\downarrow$ better); bold marks the best score in each column, excluding the oracle upper bound. $\alpha^*$ is the best fixed $\alpha$ per architecture: 0.4 (Inception-v3), 0.7 (ResNet-50), 0.8 (VGG-16), 0.1 (MaxViT-T), 0.8 (ViT-B/16). \textbf{Selected $\alpha$} is a deployable per-image choice made by a cheap 2-step/5-probe insertion probe with no access to ground truth, at $2.87\times$ the cost of AttEXplore; \textit{oracle $\alpha$} is a non-deployable upper bound --- the per-image maximum Insertion and minimum Deletion achievable in the full 11-point sweep\newedit{, each chosen independently, so the two may come from different $\alpha$ for the same image;} unobtainable without the full evaluation score itself --- reported only to bound the remaining headroom. FAMPE (all-pass) is an ablation with the frequency split disabled ($\alpha=1$, low-pass mask forced to all-ones) and reproduces AttEXplore on every architecture and both metrics, indicating that the gains are not explained by other differences between the two implementations.}

\label{tab:overall}
\end{table*}

\begin{table*}[!t]
\centering
\color{neweditcolor}
\begin{adjustbox}{max width=\textwidth}
\begin{tabular}{lcccccc}
\toprule
\textbf{Method} & \textbf{Inception-v3} & \textbf{ResNet-50} & \textbf{VGG-16} & \textbf{MaxViT-T} & \textbf{ViT-B/16} & \textbf{Mean} \\
\midrule
\multicolumn{7}{l}{\textit{Pointing game} ($\uparrow$ better)} \\
\textit{Random} & \textit{0.4930} & \textit{0.4920} & \textit{0.4990} & \textit{0.5000} & \textit{0.4920} & \textit{0.4952} \\
AttEXplore & \textbf{0.8940} & 0.7880 & \textbf{0.7850} & 0.6440 & 0.6570 & 0.7536 \\
\textbf{FAMPE ($\alpha^*$ fixed)} & 0.8720 & 0.7950 & 0.7780 & 0.6920 & 0.6810 & 0.7636 \\
\textbf{FAMPE (selected $\alpha$)} & 0.8710 & \textbf{0.8020} & 0.7790 & \textbf{0.7250} & \textbf{0.6950} & \textbf{0.7744} \\
\midrule
\multicolumn{7}{l}{\textit{Energy ratio} ($\uparrow$ better)} \\
\textit{Random} & \textit{0.4905} & \textit{0.4920} & \textit{0.4920} & \textit{0.4920} & \textit{0.4920} & \textit{0.4917} \\
AttEXplore & \textbf{0.7655} & 0.7083 & \textbf{0.6940} & 0.5771 & 0.6655 & 0.6821 \\
\textbf{FAMPE ($\alpha^*$ fixed)} & 0.7651 & 0.7141 & 0.6869 & 0.6215 & 0.6623 & 0.6900 \\
\textbf{FAMPE (selected $\alpha$)} & 0.7570 & \textbf{0.7152} & 0.6923 & \textbf{0.6554} & \textbf{0.6669} & \textbf{0.6974} \\
\bottomrule
\end{tabular}
\end{adjustbox}
\caption{\newedit{Localization against ImageNet ground-truth bounding boxes ($n=1000$ images per architecture; higher is better for both metrics). Pointing game: fraction of images whose attribution peak falls inside a box. Energy ratio: fraction of attribution mass inside the boxes. Random: a random attribution map, shown as the chance level. Mean: average over the five architectures. Bold marks the best score in each column, excluding Random; Table~\ref{tab:loc_diff} gives the significance of the differences.}}
\label{tab:loc}
\end{table*}

\begin{table*}[!t]
\centering
\color{neweditcolor}
\begin{adjustbox}{max width=\textwidth}
\begin{tabular}{lcccccc}
\toprule
\textbf{Comparison} & \textbf{Inception-v3} & \textbf{ResNet-50} & \textbf{VGG-16} & \textbf{MaxViT-T} & \textbf{ViT-B/16} & \textbf{Mean} \\
\midrule
\multicolumn{7}{l}{\textit{Pointing game} ($\uparrow$ better; positive difference = first method better)} \\
FAMPE ($\alpha^*$ fixed) $-$ AttEXplore & $-0.0220^{*}$ & $+0.0070$ & $-0.0070$ & $+0.0480^{*}$ & $+0.0240$ & $+0.0100$ \\
 & \scriptsize[$-0.0390,-0.0060$] & \scriptsize[$-0.0160,+0.0300$] & \scriptsize[$-0.0290,+0.0150$] & \scriptsize[$+0.0110,+0.0860$] & \scriptsize[$-0.0070,+0.0560$] & \scriptsize--- \\
\addlinespace[2pt]
FAMPE (selected $\alpha$) $-$ AttEXplore & $-0.0230^{*}$ & $+0.0140$ & $-0.0060$ & $+0.0810^{*}$ & $+0.0380^{*}$ & $+0.0208$ \\
 & \scriptsize[$-0.0410,-0.0050$] & \scriptsize[$-0.0100,+0.0380$] & \scriptsize[$-0.0270,+0.0150$] & \scriptsize[$+0.0450,+0.1170$] & \scriptsize[$+0.0060,+0.0700$] & \scriptsize--- \\
\addlinespace[2pt]
FAMPE (selected $\alpha$) $-$ FAMPE ($\alpha^*$ fixed) & $-0.0010$ & $+0.0070$ & $+0.0010$ & $+0.0330^{*}$ & $+0.0140$ & $+0.0108$ \\
 & \scriptsize[$-0.0180,+0.0160$] & \scriptsize[$-0.0130,+0.0280$] & \scriptsize[$-0.0190,+0.0200$] & \scriptsize[$+0.0040,+0.0620$] & \scriptsize[$-0.0160,+0.0440$] & \scriptsize--- \\
\midrule
\multicolumn{7}{l}{\textit{Energy ratio} ($\uparrow$ better; positive difference = first method better)} \\
FAMPE ($\alpha^*$ fixed) $-$ AttEXplore & $-0.0004$ & $+0.0059^{*}$ & $-0.0070^{*}$ & $+0.0444^{*}$ & $-0.0031$ & $+0.0080$ \\
 & \scriptsize[$-0.0022,+0.0014$] & \scriptsize[$+0.0017,+0.0101$] & \scriptsize[$-0.0116,-0.0024$] & \scriptsize[$+0.0334,+0.0558$] & \scriptsize[$-0.0111,+0.0051$] & \scriptsize--- \\
\addlinespace[2pt]
FAMPE (selected $\alpha$) $-$ AttEXplore & $-0.0086^{*}$ & $+0.0070^{*}$ & $-0.0017$ & $+0.0783^{*}$ & $+0.0014$ & $+0.0153$ \\
 & \scriptsize[$-0.0131,-0.0041$] & \scriptsize[$+0.0016,+0.0123$] & \scriptsize[$-0.0068,+0.0034$] & \scriptsize[$+0.0654,+0.0912$] & \scriptsize[$-0.0079,+0.0106$] & \scriptsize--- \\
\addlinespace[2pt]
FAMPE (selected $\alpha$) $-$ FAMPE ($\alpha^*$ fixed) & $-0.0082^{*}$ & $+0.0011$ & $+0.0053^{*}$ & $+0.0338^{*}$ & $+0.0046$ & $+0.0073$ \\
 & \scriptsize[$-0.0126,-0.0038$] & \scriptsize[$-0.0034,+0.0058$] & \scriptsize[$+0.0006,+0.0102$] & \scriptsize[$+0.0251,+0.0426$] & \scriptsize[$-0.0037,+0.0129$] & \scriptsize--- \\
\bottomrule
\end{tabular}
\end{adjustbox}
\caption{\newedit{Paired differences in localization between the methods of Table~\ref{tab:loc}. Each entry is the difference in score between the two named methods; since higher scores are better for both metrics, a positive value means the first method localizes better. The first two rows of each block compare the two FAMPE configurations with AttEXplore; the third row isolates the effect of per-image $\alpha$ selection by comparing it with the fixed $\alpha^*$. Below each difference is its 95\% CI from a paired bootstrap over images (10{,}000 resamples, seed 42); * marks differences whose CI excludes zero. Mean: average of the five per-architecture differences, without a CI because the bootstrap is run per architecture.}}
\label{tab:loc_diff}
\end{table*}

\begin{figure*}[!t]
  \centering
  \includegraphics[width=\linewidth]{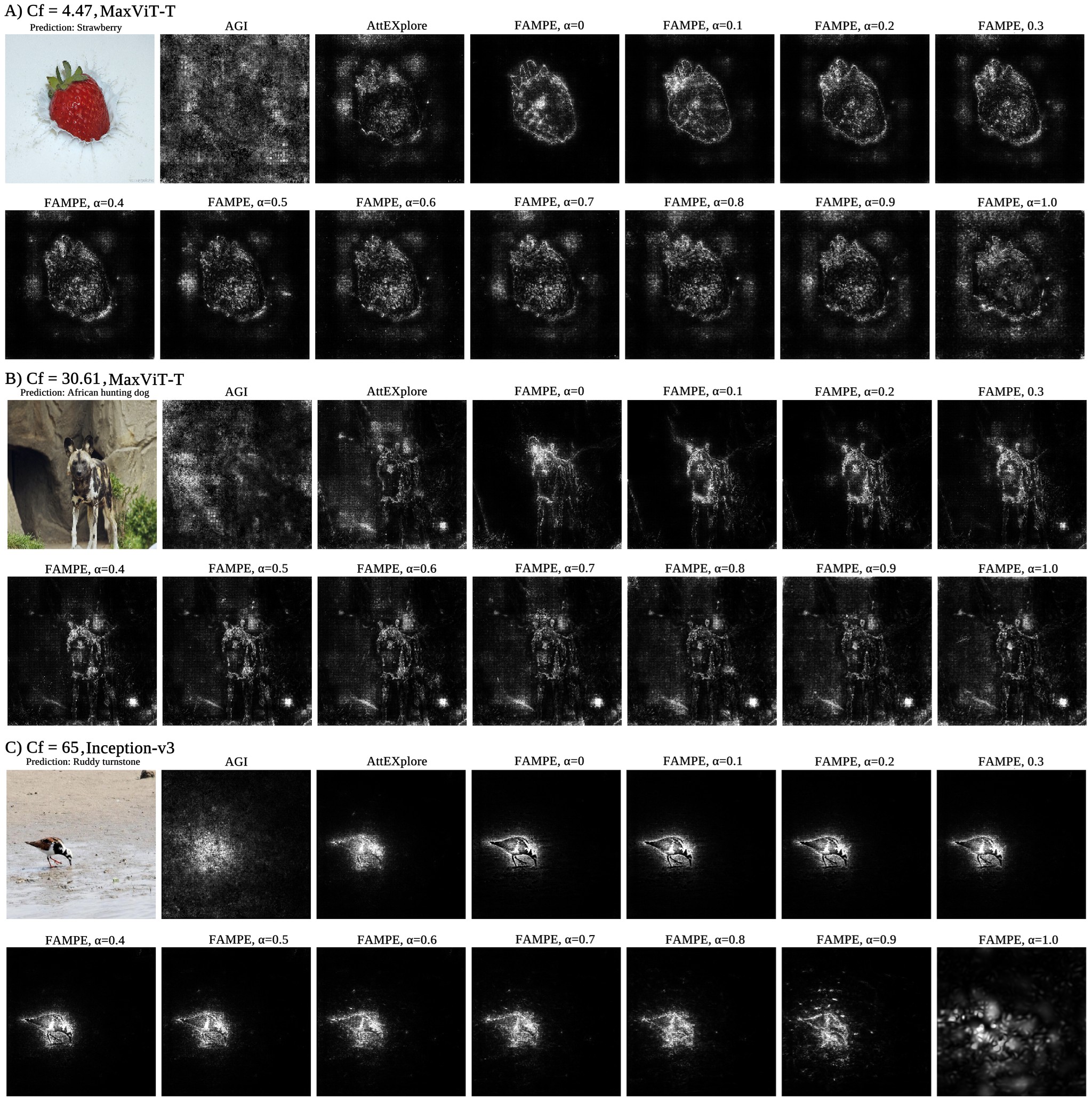}
  \caption{Qualitative comparison of attribution maps from FAMPE, AGI, and AttEXplore on three ImageNet samples of varying spectral complexity: (A) a strawberry on MaxViT-T (low-frequency content, small $c_f$), (B) an African hunting dog on MaxViT-T (mixed mid- to high-frequency content, intermediate $c_f$), and (C) a ruddy turnstone on Inception-v3 (high-frequency content, large $c_f$). For each panel, AGI and AttEXplore maps are followed by 11 FAMPE maps with $\alpha$ from 0 (purely high-frequency perturbation) to 1 (purely low-frequency perturbation). {On these two architectures ($\alpha^*=0.1$ and $0.4$), lower-$\alpha$ FAMPE maps place mass more tightly on the target object than the baselines.}}
  \label{fig3}
\end{figure*}

\noindent \textbf{Qualitative comparison.} \hspace{0.5em} {In the examples of Fig.~\ref{fig3}, all from MaxViT-T and Inception-v3 (the two architectures with the lowest $\alpha^*$), low-$\alpha$ FAMPE maps ($\alpha \leq 0.4$) place mass more tightly on the target object than AGI and AttEXplore. As $\alpha$ increases, the maps broaden and lose contrast, most visibly on Inception-v3, where $\alpha = 1.0$ no longer resolves the object outline. The three samples vary in spectral complexity (further results in \ref{sec:appendix_attribution}): a strawberry on MaxViT-T with a small $c_f$ (low-frequency-dominated), an African hunting dog on MaxViT-T with intermediate $c_f$ (textured fur), and a ruddy turnstone on Inception-v3 with the largest $c_f$ (high-frequency content across plumage and background). On VGG-16, ResNet-50 and ViT-B/16, the trend reverses and a high $\alpha$ gives the best Insertion (Table~\ref{tab:alpha_tab}).}

{
\noindent \textbf{Fidelity of the $\alpha$ probe.} \hspace{0.5em} The probe does not reliably find the single best $\alpha$ of an image, but it reliably separates better from worse regions of the sweep, which is what an argmax selector needs. Its choice matches the best $\alpha$ of the full 11-point sweep for 26\% to 53\% of the images, depending on the architecture (chance: $1/11 \approx 9\%$), and the best $\alpha$ is among its top three candidates for 55\% to 84\%. Probe and full-sweep Insertion scores correlate at $r \approx 0.97$ over all (image, $\alpha$) pairs and at $r \approx 0.74$ on average within individual images.
}

\subsection{Ablation Studies}
\label{sec:ablation}

\nopagebreak
\noindent \textbf{Effect of $\alpha$.} \hspace{0.5em}
The choice of $\alpha$ matters on every architecture: the worst $\alpha$ in the sweep loses $19\%$ (MaxViT-T) to $36\%$ (ViT-B/16) of the best mean Insertion Score (Table~\ref{tab:alpha_tab}). The best $\alpha$ depends on the architecture and does not follow a CNN-versus-Transformer split. MaxViT-T ($\alpha^*=0.1$) and Inception-v3 ($\alpha^*=0.4$) favor a low $\alpha$, which emphasizes the high-frequency band, whereas ResNet-50 ($\alpha^*=0.7$), VGG-16 ($0.8$) and ViT-B/16 ($0.8$) favor a high $\alpha$, which emphasizes the low-frequency band. On average, perturbing only one band is never optimal: $\alpha^*$ lies strictly between 0 and 1 on every architecture. Finally, the best $\alpha$ also varies from image to image: on no architecture is a single $\alpha$ optimal for more than $26\%$ of the images. Note that $\alpha=0.5$ weights both bands equally, so the noise-free input is scaled by one half; its Insertion Score is almost the same as AttEXplore's (within $0.021$) on every architecture. These results suggest a benefit beyond uniform attenuation alone, but do not isolate the contributions of signal weighting and stochastic perturbation.

\begin{table*}[!t]
\centering

\begin{adjustbox}{max width=\textwidth}
\begin{tabular}{lcccccccccc}
\toprule
\multirow{2}{*}{\textbf{$\alpha$}}
  & \multicolumn{2}{c}{\textbf{Inception-v3}}
  & \multicolumn{2}{c}{\textbf{ResNet-50}}
  & \multicolumn{2}{c}{\textbf{VGG-16}}
  & \multicolumn{2}{c}{\textbf{MaxViT-T}}
  & \multicolumn{2}{c}{\textbf{ViT-B/16}} \\
\cmidrule(lr){2-3} \cmidrule(lr){4-5} \cmidrule(lr){6-7} \cmidrule(lr){8-9} \cmidrule(lr){10-11}
 & \textbf{Ins.} & \textbf{Del.}
 & \textbf{Ins.} & \textbf{Del.}
 & \textbf{Ins.} & \textbf{Del.}
 & \textbf{Ins.} & \textbf{Del.}
 & \textbf{Ins.} & \textbf{Del.} \\
\midrule
0 & 0.4692 (11) & 0.1242 (7) & 0.2527 (5) & 0.1436 (6) & 0.2181 (8) & 0.1083 (11) & 0.4294 (26) & 0.2034 (26) & 0.3195 (5) & 0.1997 (6) \\
0.1 & 0.5114 (13) & 0.1099 (9) & 0.3015 (8) & 0.1190 (10) & 0.2584 (10) & 0.0916 (12) & \textbf{0.4419 (19)} & 0.1974 (19) & 0.3699 (4) & 0.1469 (15) \\
0.2 & 0.5272 (13) & 0.1027 (10) & 0.3258 (9) & 0.1093 (9) & 0.2669 (7) & 0.0900 (9) & 0.4294 (12) & 0.2145 (9) & 0.4211 (7) & 0.1333 (17) \\
0.3 & 0.5300 (10) & 0.1006 (8) & 0.3411 (9) & 0.1055 (9) & 0.2710 (8) & 0.0893 (10) & 0.4225 (8) & 0.2226 (6) & 0.4396 (7) & 0.1343 (14) \\
0.4 & \textbf{0.5314 (9)} & 0.0990 (6) & 0.3463 (8) & 0.1042 (9) & 0.2759 (6) & 0.0891 (7) & 0.4148 (7) & 0.2282 (5) & 0.4533 (6) & 0.1422 (10) \\
0.5 & 0.5298 (9) & 0.0988 (7) & 0.3481 (8) & 0.1028 (9) & 0.2787 (7) & 0.0900 (4) & 0.4039 (5) & 0.2377 (4) & 0.4674 (7) & 0.1461 (8) \\
0.6 & 0.5264 (7) & 0.0985 (6) & 0.3525 (8) & 0.1025 (7) & 0.2834 (7) & 0.0879 (6) & 0.3958 (4) & 0.2443 (3) & 0.4825 (9) & 0.1516 (7) \\
0.7 & 0.5210 (7) & 0.0984 (7) & \textbf{0.3575 (9)} & 0.1020 (8) & 0.2895 (9) & 0.0860 (6) & 0.3879 (5) & 0.2454 (4) & 0.4979 (12) & 0.1565 (7) \\
0.8 & 0.5106 (6) & 0.0985 (8) & 0.3572 (8) & 0.1016 (10) & \textbf{0.2909 (10)} & 0.0837 (8) & 0.3836 (5) & 0.2447 (5) & \textbf{0.5028 (15)} & 0.1613 (4) \\
0.9 & 0.4957 (9) & 0.0984 (10) & 0.3492 (12) & 0.1038 (9) & 0.2896 (13) & 0.0826 (10) & 0.3728 (5) & 0.2397 (7) & 0.4853 (14) & 0.1674 (5) \\
1.0 & 0.4057 (6) & 0.1045 (23) & 0.3344 (15) & 0.1047 (14) & 0.2763 (15) & 0.0837 (17) & 0.3571 (5) & 0.2312 (12) & 0.4694 (13) & 0.1751 (7) \\
\bottomrule
\end{tabular}
\end{adjustbox}
\caption{Ablation study on the hyperparameter $\alpha$ across five architectures: mean Insertion Score (Ins., $\uparrow$ better) and Deletion Score (Del., $\downarrow$ better). The number in parentheses is the percentage of images whose per-image maximum Insertion Score (Ins.) or minimum Deletion Score (Del.) is attained at this $\alpha$ within the 11-point sweep, rounded to the nearest integer; each column sums to 100 up to rounding. Bold marks the single highest mean Insertion Score in each network's column, matching the $\alpha^*$ used as the fixed setting in Table~\ref{tab:overall}. $n=1000$ per network, the same image set used throughout.}

\label{tab:alpha_tab}
\end{table*}

\begin{figure*}[!t]
  \centering
  \includegraphics[width=0.85\linewidth]{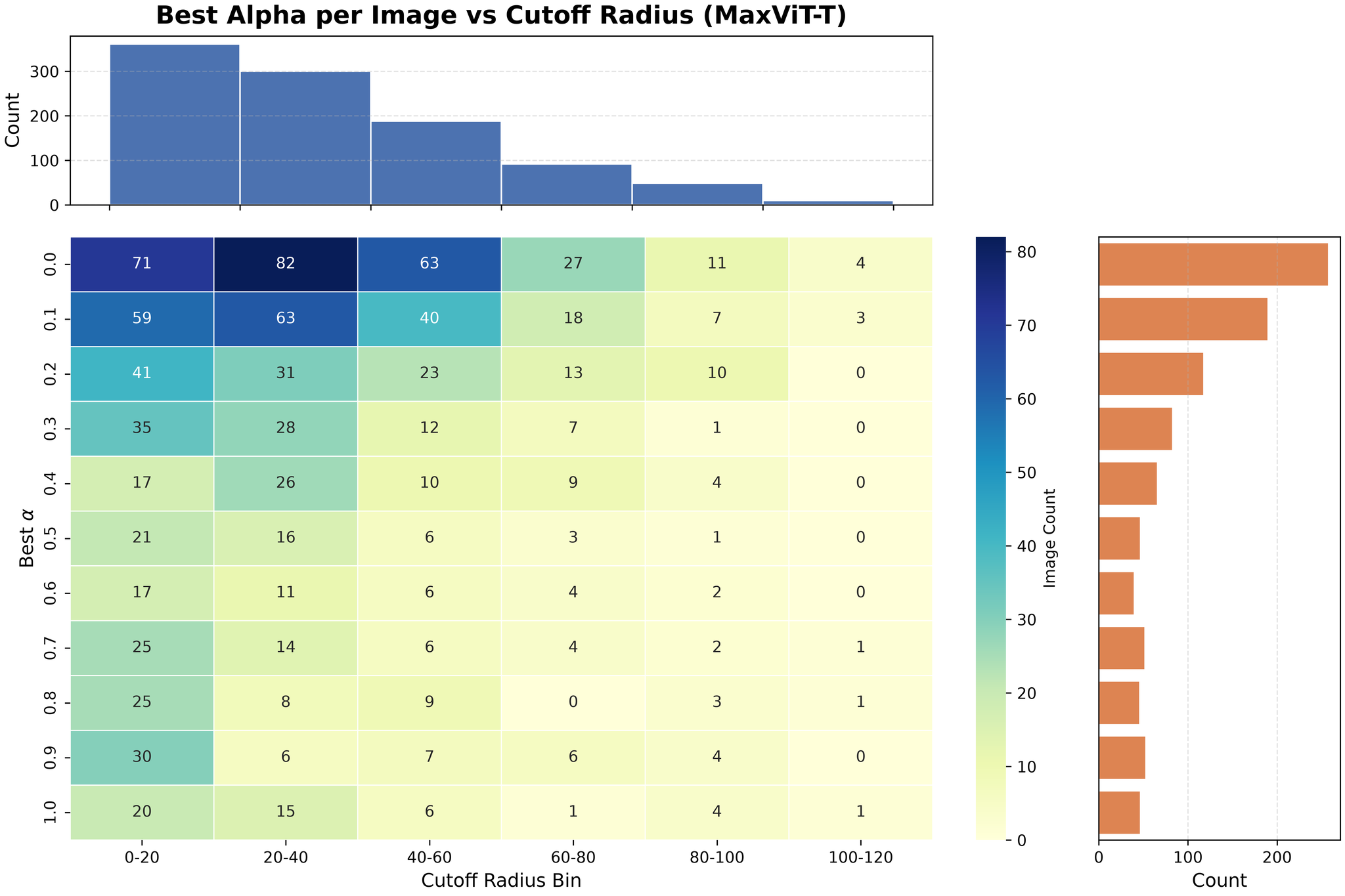}
\caption{Joint distribution of the per-image cutoff $c_f$ and the best-$\alpha$ value (maximizing Insertion Score) across 1,000 ImageNet samples on MaxViT-T; $c_f$ is recomputed with the estimator FAMPE uses at attribution time, on the preprocessed image at the network's input size. The central heatmap shows image counts per $(c_f, \alpha)$ bin (darker = higher density), with marginal histograms of cutoff radii (top) and best-$\alpha$ values (right). Both marginals are skewed --- most images have a small $c_f$, and most are best served by a low $\alpha$ --- but the two are close to independent (upper-left quadrant: 410 images observed vs.\ 429 expected; Cram\'er's $V = 0.12$).}
  \label{fig4}
\end{figure*}

\noindent \textbf{Joint structure of $c_f$ and best-$\alpha$.} \hspace{0.5em} An image's best $\alpha$ is nearly independent of its spectral cutoff $c_f$ (Fig.~\ref{fig4}, MaxViT-T). The heatmap is densest in the upper-left, where images have a small cutoff ($c_f < 40$) and a low best $\alpha$ ($\alpha < 0.4$), but this only reflects the two skewed marginals: 661 of 1{,}000 images have $c_f < 40$, and 649 are best served by $\alpha < 0.4$. Under independence, 429 images would fall in this quadrant; we observe 410. A $\chi^2$ test detects a dependence ($p = 0.014$), but it is negligible (Cram\'er's $V = 0.12$) and runs opposite to the intuitive reading: images best served by a low $\alpha$ have a slightly higher median cutoff (29.2 vs.\ 22.5; Spearman $\rho = -0.156$).

\subsection{Computational Cost}
\label{sec:cost}
{\nopagebreak FAMPE at a fixed $\alpha$ costs essentially the same as AttEXplore, and per-image selection still costs less than AGI (Table~\ref{tab:compute}). With a fixed $\alpha$, both methods use an identical budget by construction (10 attack steps with $N=20$: 200 backward and 200 forward passes) and differ by less than 3\% in wall-clock time. Selecting $\alpha$ adds the truncated 11-candidate probe (110 backward and 231 forward passes) to one full-cost run, for $2.87\times$ the wall-clock cost of AttEXplore. This is still $0.78\times$ the cost of AGI, which requires four times as many backward passes as a single FAMPE run.}

\begin{table}[!t]
\centering

\begin{adjustbox}{max width=\columnwidth}
\begin{tabular}{lcccc}
\toprule
\textbf{Method} & \textbf{\shortstack{Backward\\passes}} & \textbf{\shortstack{Forward\\passes}} & \textbf{\shortstack{s/image\\(VGG-16)}} & \textbf{\shortstack{Relative\\cost}} \\
\midrule
AGI & 800 & 401 & 1.073 & 3.68$\times$ \\
AttEXplore & 200 & 200 & 0.292 & 1.00$\times$ \\
FAMPE ($\alpha^*$ fixed) & 200 & 200 & 0.300 & 1.03$\times$ \\
FAMPE (selected $\alpha$) & 310 & 431 & 0.837 & 2.87$\times$ \\
\bottomrule
\end{tabular}
\end{adjustbox}
\caption{Per-sample computational cost. Backward and forward pass counts follow directly from each method's protocol (Parameters, above) and are the same on every architecture; wall-clock is measured on VGG-16 on an idle RTX 4090 and reported relative to AttEXplore. FAMPE (selected $\alpha$) totals the 11-candidate truncated probe (110 backward / 231 forward passes: 110 gradient evaluations plus 121 forward passes to score the coarse maps) and one full-cost run (200/200) at the chosen $\alpha$.}
\label{tab:compute}
\end{table}

\section{Discussion}
\label{sec:discussion}
{\nopagebreak
\noindent \textbf{The useful band is a property of the model.} \hspace{0.5em} The central finding of this work is that the frequency band an attribution method perturbs matters, and that the best band depends on the model being explained. FORGrad~\citep{muzellec2024saliency} found a related dependence for white-box methods, whose best low-pass cutoff on the gradient varies with the architecture. Here the dependence also concerns the direction of the weighting: MaxViT-T and Inception-v3 favor the high-frequency band, a setting that a low-pass filter cannot represent. The same images yield different $\alpha^*$ on different architectures, with no split between CNNs and Transformers, and an image's best $\alpha$ is nearly independent of its spectral cutoff (Sec.~\ref{sec:ablation}). An all-pass perturbation, as used by AttEXplore, cannot adapt to this, because it perturbs all bands alike regardless of which one matters for the model. FAMPE makes the band a tunable choice. Since the right choice follows neither the architecture family nor the image's spectral cutoff, it has to be measured against the model: once per architecture (fixed $\alpha^*$) or per image (Sec.~\ref{sec:selection}). The $\alpha$ sweep may thus also serve to indicate which band is most informative about a given model's decisions.

\noindent \textbf{Insertion gains do not guarantee better localization.} \hspace{0.5em} FAMPE raises Insertion in two ways, which behave differently under the localization check. The complete frequency-weighted procedure, at a fixed $\alpha^*$, improves localization on some architectures, most clearly on MaxViT-T. Per-image selection, which directly maximizes a truncated Insertion score, raises Insertion on every architecture but improves localization in only three of ten comparisons, and on Inception-v3 it even lowers the energy ratio. Optimizing an attribution method directly against Insertion can therefore raise the score without producing a better-localized map, in line with the observation of Yavari et al.~\citep{Yavari2026} that Insertion rankings can disagree with expert assessment. This does not make Insertion wrong. Localization measures agreement with the annotated object, whereas Insertion measures agreement with the model, and a faithful map may legitimately highlight context outside the box. It does mean that attribution methods whose hyperparameters are tuned on Insertion should also be validated with an independent measure.

\noindent \textbf{FAMPE helps most where the all-pass baseline is weakest.} \hspace{0.5em} The gains are largest on MaxViT-T, where AttEXplore localizes worst, and, at the fixed $\alpha^*$, absent or negative on Inception-v3, where AttEXplore is already best on all three metrics. On such a model, the fixed configuration offers no gain, only the risk of a poorly chosen $\alpha$, which costs up to $24\%$ of Insertion; per-image selection still raises Insertion there, but lowers localization. Headroom is not the whole story, however: ViT-B/16 also starts from a low localization baseline, yet at the fixed $\alpha^*$ it gains significantly only on Insertion.

\noindent \textbf{Which configuration to use.} \hspace{0.5em} We recommend the fixed $\alpha^*$ as the default. It costs essentially the same as AttEXplore, improves Insertion on four of five architectures, and its gains do not rely on per-image optimization against Insertion; the only extra cost is a one-time sweep per architecture to determine $\alpha^*$. Per-image selection is an optional, Insertion-specific upgrade: it raises Insertion on every architecture at $2.87\times$ the cost of AttEXplore, but its benefit should be validated against an independent measure before it is relied on in a new setting.

\noindent \textbf{Scope of the comparison.} \hspace{0.5em} Our claim is scoped to a mechanism, not to the field. We test whether separating the spectrum into independently perturbed high- and low-frequency bands improves attribution within the model parameter exploration framework. Answering this requires a comparison in which the separation is the only variable. AttEXplore is therefore the primary comparator: it shares FAMPE's exploration backbone and settings ($\eta$, $N$, $\sigma$, $\epsilon$ and the number of attack iterations) \newedit{but lacks FAMPE's frequency awareness}. Our absolute scores differ from those published for AttEXplore \citep{zhu2024attexplore}. Such scores depend on the image sample, the step grid of the Insertion and Deletion curves, their baseline image and the preprocessing, none of which are standardized across papers. All comparisons here are therefore made within one protocol, on identical images. The all-pass control (Table~\ref{tab:overall}; $\alpha=1$ with the low-pass mask forced to all ones) reproduces AttEXplore's Insertion and Deletion scores to within $\pm0.0026$ on every architecture, indicating that, within our implementation, the differences we report are associated with the frequency-weighted split rather than with other implementation differences.

\noindent \textbf{Limitations.} \hspace{0.5em} At the fixed $\alpha^*$, the localization benefit of FAMPE is architecture-dependent: it is clear on MaxViT-T, small on ResNet-50 (energy ratio only), and absent or negative on the other three architectures. The $\alpha$ selector is scored with a truncated form of Insertion, so its benefit is largely specific to that metric; on Inception-v3 it even lowers localization relative to AttEXplore. Finally, our evaluation is confined to single-label ImageNet classification; whether the spectral behavior we report transfers to other tasks, such as detection or segmentation, or to other data domains remains untested.
}

\section{Conclusion}
{\nopagebreak We \newedit{introduce} FAMPE, an attribution method that perturbs the low- and high-frequency bands of the input independently instead of perturbing the whole spectrum alike. A per-image, energy-based cutoff separates the two bands, and a single parameter $\alpha$ sets their weights \newedit{at essentially no extra cost. This} makes attributions more faithful to the model than those of the strongest baseline, AttEXplore, on most architectures. The results carry two implications beyond FAMPE. First, the frequency content of the perturbation is a design choice that shapes the explanation, and the right choice depends on the model, so it should be measured against the model rather than assumed. Second, gains in Insertion do not guarantee better-localized explanations, so attribution methods tuned on Insertion should also be validated with an independent measure. This work demonstrates the potential of frequency-aware perturbations for attribution, extending the spectral perspective on attribution~\citep{muzellec2024saliency} from the gradient to the perturbation.}

\section*{CRediT authorship contribution statement}
\newedit{\textbf{Ali Yavari:} Conceptualization, Methodology, Software, Validation, Formal analysis, Investigation, Data curation, Visualization, Writing -- original draft, Writing -- review \& editing. \textbf{Alireza Mohamadi:} Investigation. \textbf{Elham Beydaghi:} Writing -- review \& editing. \textbf{Philipp Seeb\"{o}ck:} Conceptualization, Methodology, Investigation, Formal analysis, Validation, Writing -- review \& editing. \textbf{Rainer A. Leitgeb:} Conceptualization, Methodology, Investigation, Formal analysis, Validation, Project administration, Supervision.}

\section*{Declaration of competing interest}
\newedit{The authors declare that they have no known competing financial interests or personal relationships that could have appeared to influence the work reported in this paper.}

\section*{Declaration of generative AI and AI-assisted technologies in the manuscript preparation process}
\newedit{During the preparation of this work, the authors used Claude (Anthropic) only to improve the language and readability of the manuscript; it was not used to generate scientific content or references. All changes were carefully reviewed and revised by the authors, who take full responsibility for the content of the published article.}

\section*{Funding}
\newedit{This research did not receive any specific grant from funding agencies in the public, commercial, or not-for-profit sectors.}

\section*{Data availability}
\newedit{The code will be made publicly available upon acceptance. The ImageNet data are available from the original ImageNet source~\citep{deng2009imagenet}, subject to its terms of use.}

\appendix

\section{Axiomatic Properties}
\label{sec:appendix_axioms}

An attribution method satisfies \textit{Sensitivity} when, if an input and its baseline differ in only one feature that leads to different predictions, that feature is assigned a non-zero attribution. An attribution method satisfies \textit{Implementation Invariance} when two neural networks producing identical outputs for all inputs yield identical attributions. FAMPE satisfies Implementation Invariance when both networks are run with the same random draws in Eq.~\eqref{eq:main_eq} and are differentiable at the evaluated points, because it then depends on the network only through its outputs and input gradients, which coincide for functionally equivalent networks. Under the assumptions stated below, an approximate completeness relation follows: the attribution summed over input features approximately equals the change in loss along the exploration path. This relation does not establish Sensitivity, which concerns an input and a baseline that differ in a single feature, whereas the endpoint of FAMPE's exploration generally differs from the input in many features.

\noindent\textit{Approximate completeness.} FAMPE changes only how the perturbed variants $x^t_{\omega_i}$ are generated (Eq.~\eqref{eq:main_eq}); the integration form of Eq.~\eqref{eq:path_eq} and the update rule of Eqs.~\eqref{eq:minor_eq_one}--\eqref{eq:minor_eq_two} are those of AttEXplore~\citep{zhu2024attexplore}, so the argument given there carries over, and we restate it in our notation. We show that the accumulated attribution, summed over the input features $j$, approximately equals the change in loss between the start $x^0$ and the endpoint $x^T$ of the exploration path,
\begin{equation}
\sum_j A_j\approx L(x^T)-L(x^0).
\label{eq:A_loss_def}
\end{equation}
Along the path, the loss at each iterate is related to the loss at the previous one by a first-order expansion,
\begin{equation}
L(x^t)=L(x^{t-1})+(x^t-x^{t-1})^{\top}\frac{\partial L(x^{t-1})}{\partial x^{t-1}}+\mathcal{O}\big(\lVert x^t-x^{t-1}\rVert^2\big).
\end{equation}
Summing this expansion over all iterates and dropping the second-order remainders gives
\begin{equation}
\sum_{t=1}^{T} L(x^t)\approx\sum_{t=0}^{T-1}L(x^{t})+\sum_{t=0}^{T-1}(x^{t+1}-x^{t})^{\top}\frac{\partial L(x^{t})}{\partial x^{t}},
\end{equation}
where the right-hand side has been re-indexed by $t'=t-1$. Canceling the loss terms common to both sides leaves the endpoint loss as the starting loss plus the accumulated first-order increments,
\begin{equation}
L(x^T)\approx L(x^0)+\sum_{t=0}^{T-1}(x^{t+1}-x^{t})^{\top}\frac{\partial L(x^{t})}{\partial x^{t}}.
\label{eq:loss_increments}
\end{equation}
Summing the attribution of Eq.~\eqref{eq:path_eq} over features and comparing with Eq.~\eqref{eq:loss_increments} gives
\begin{multline}
\sum_j A_j=\sum_j\int\Delta x^t_j\,g_j(x^t)\,dt \\
\approx\sum_{t=0}^{T-1}(x^{t+1}-x^{t})^{\top}\frac{\partial L(x^{t})}{\partial x^{t}}\approx L(x^T)-L(x^0),
\end{multline}
where $\Delta x^t$ is the increment $x^{t+1}-x^t$ of Eq.~\eqref{eq:minor_eq_one}, the integral is the sum over the $T$ steps, and the first approximation replaces $g(x^t)$, the gradient averaged over the $N$ perturbed variants in Eq.~\eqref{eq:minor_eq_two}, by the gradient at the iterate, $\partial L(x^{t})/\partial x^{t}$. Completeness therefore holds only approximately: it requires this replacement to be accurate and the steps to be small enough for the first-order expansion. It does not by itself imply Sensitivity, because the axiom concerns an input and a baseline that differ in a single feature, a setting that FAMPE's exploration does not produce.

\section{Attribution Results}
\label{sec:appendix_attribution}
\setcounter{figure}{0}

\begin{figure*}[p]
  \centering
  \includegraphics[width=0.82\textwidth]{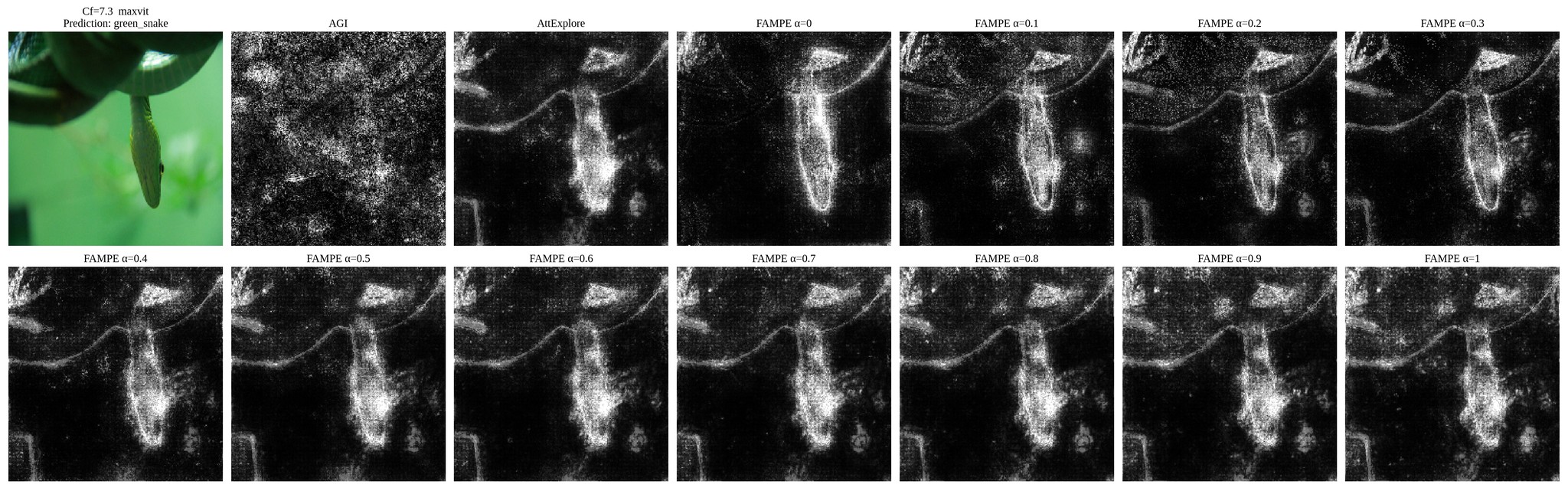}\\[2pt]
  {\small (a) Sample 1, MaxViT-T}\\[8pt]
  \includegraphics[width=0.82\textwidth]{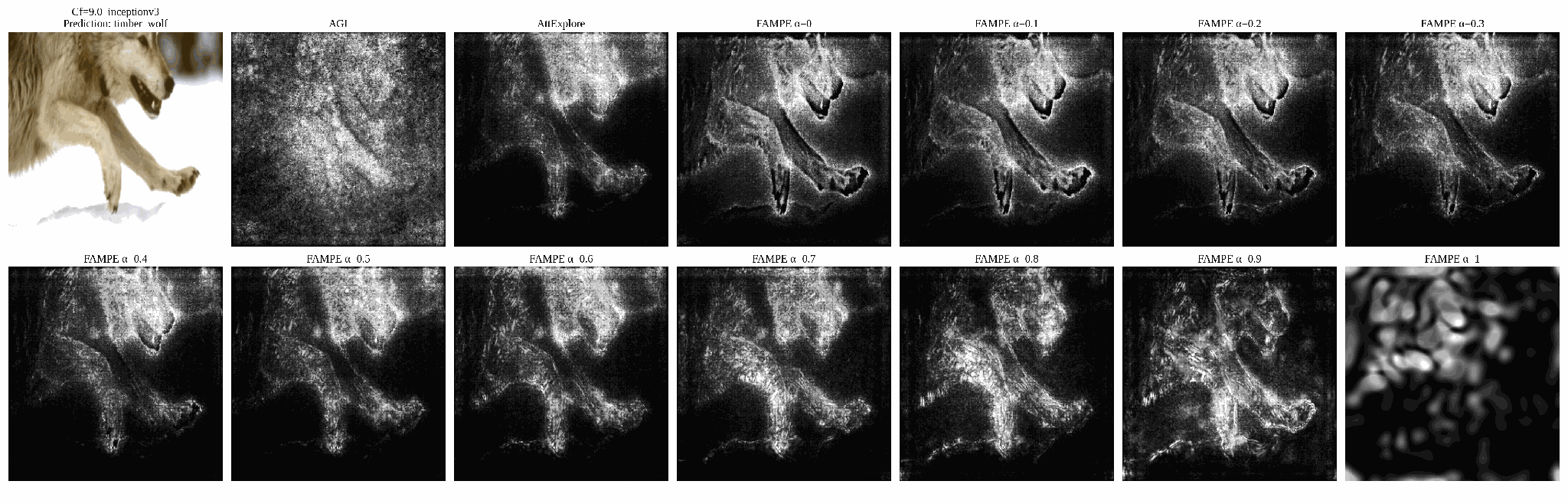}\\[2pt]
  {\small (b) Sample 2, Inception-v3}\\[8pt]
  \includegraphics[width=0.82\textwidth]{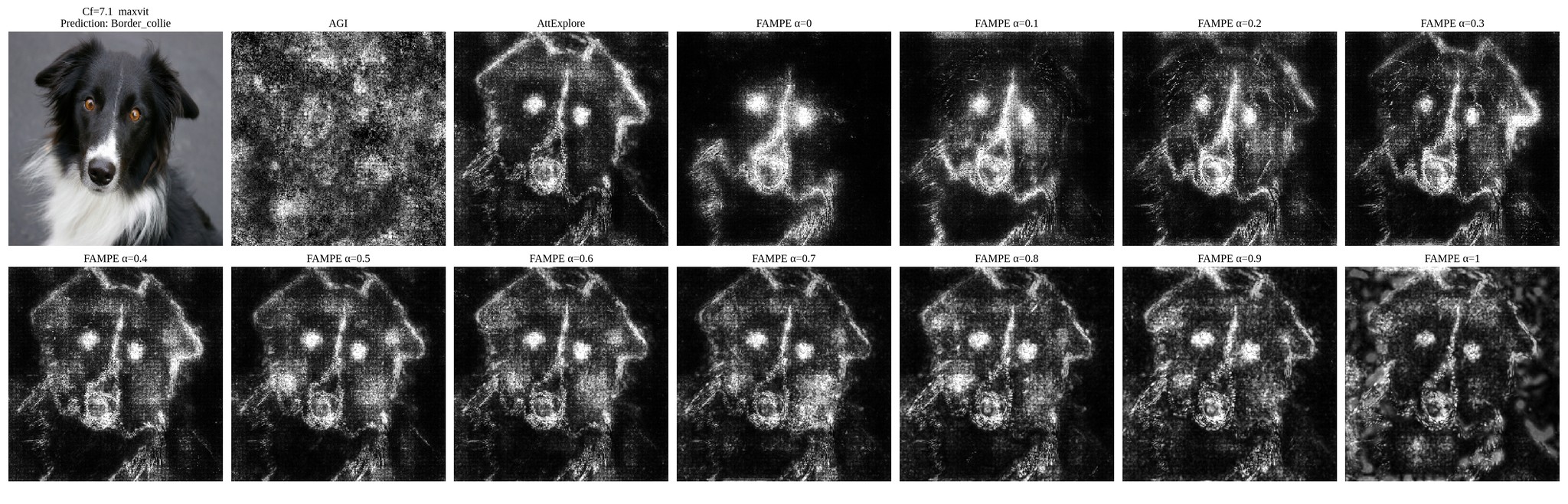}\\[2pt]
  {\small (c) Sample 3, MaxViT-T}\\[8pt]
  \includegraphics[width=0.82\textwidth]{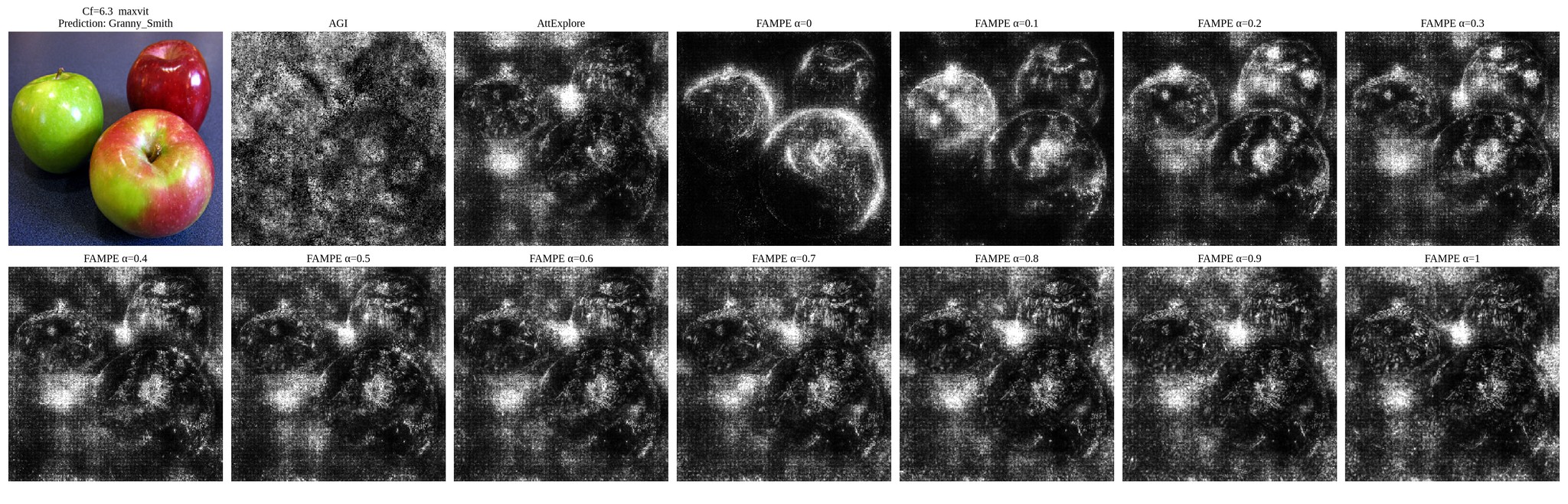}\\[2pt]
  {\small (d) Sample 4, MaxViT-T}
  \caption{\newedit{Attribution maps for low-frequency images (small $c_f$). Each panel shows the input image, the AGI and AttEXplore maps, and the FAMPE maps for $\alpha = 0, 0.1, \ldots, 1$.}}
  \label{fig:appB_low}
\end{figure*}

\begin{figure*}[p]
  \centering
  \includegraphics[width=0.82\textwidth]{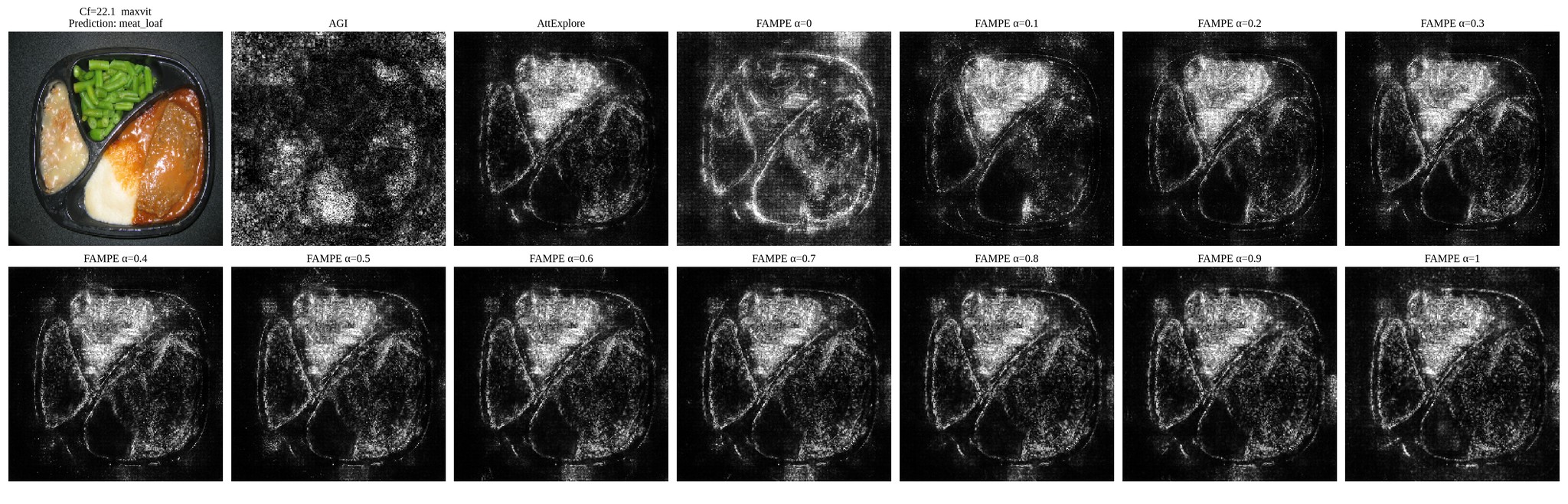}\\[2pt]
  {\small (a) Sample 5, MaxViT-T}\\[8pt]
  \includegraphics[width=0.82\textwidth]{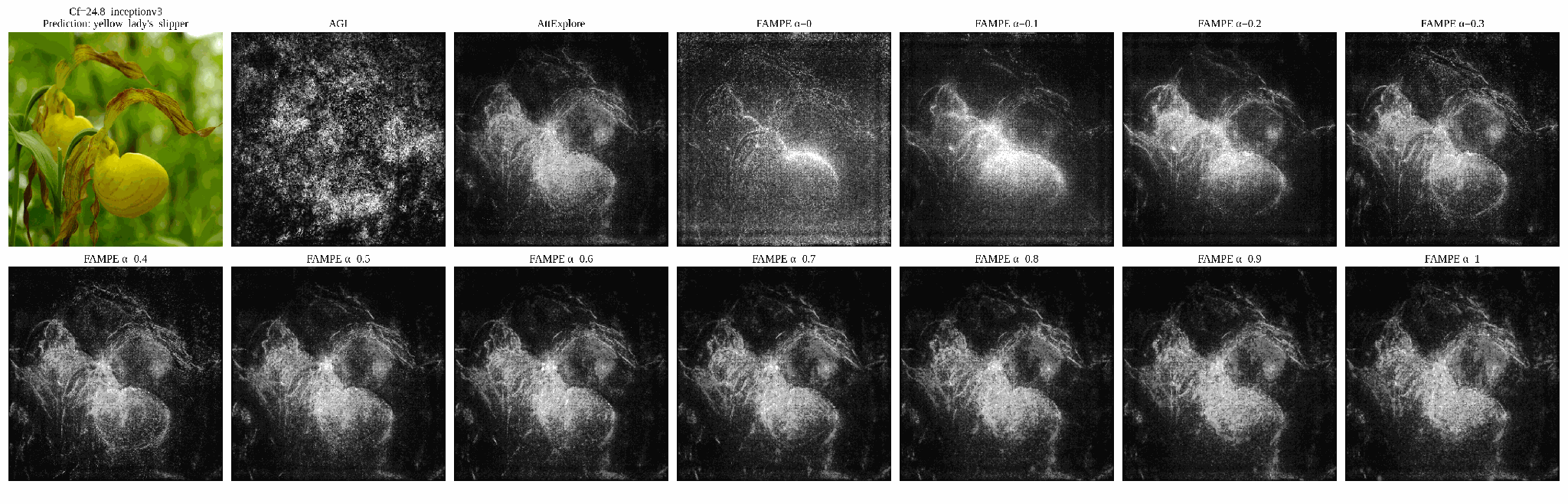}\\[2pt]
  {\small (b) Sample 6, Inception-v3}\\[8pt]
  \includegraphics[width=0.82\textwidth]{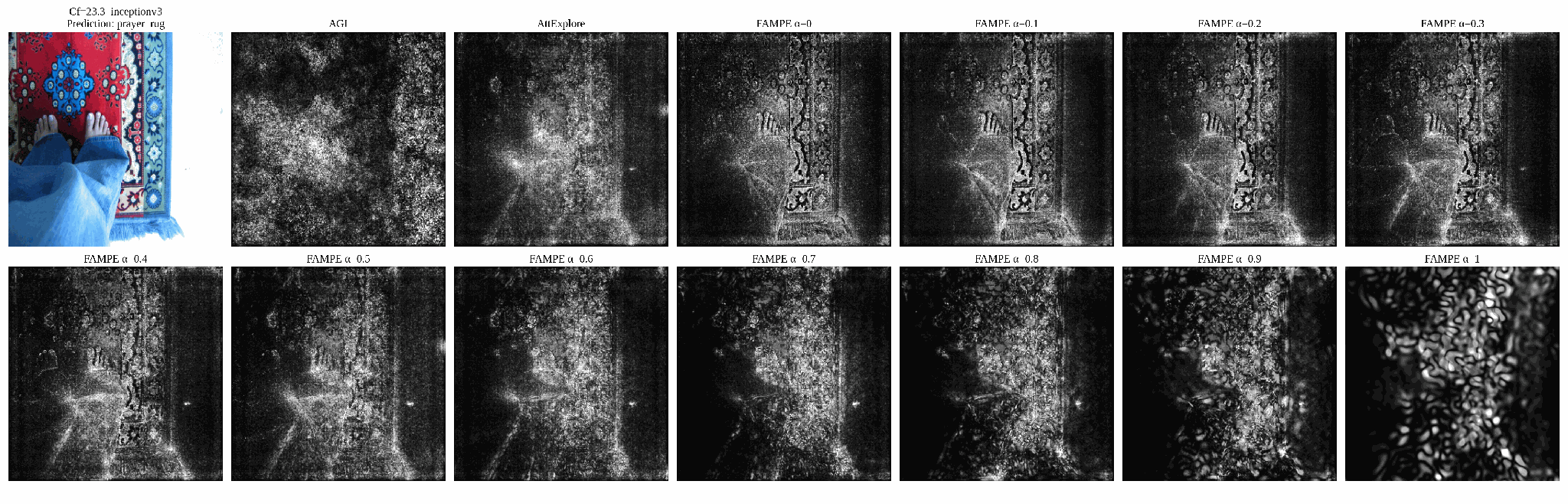}\\[2pt]
  {\small (c) Sample 7, Inception-v3}\\[8pt]
  \includegraphics[width=0.82\textwidth]{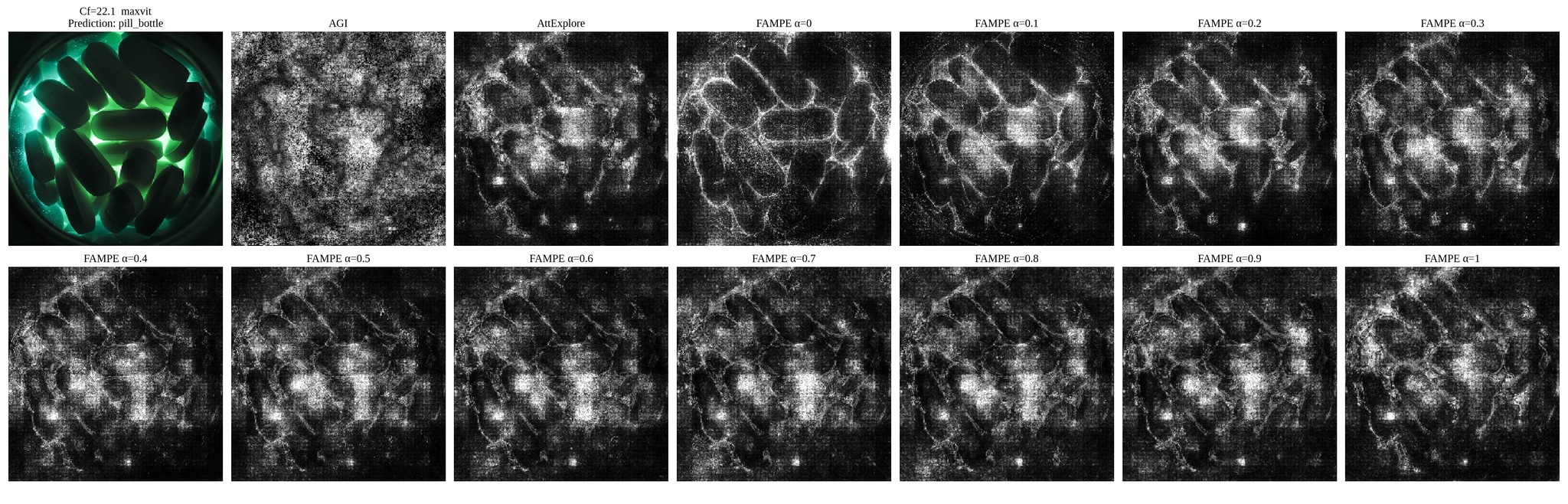}\\[2pt]
  {\small (d) Sample 8, MaxViT-T}
  \caption{\newedit{Attribution maps for middle-frequency images (intermediate $c_f$). Layout as in Fig.~\ref{fig:appB_low}.}}
  \label{fig:appB_mid}
\end{figure*}

\begin{figure*}[p]
  \centering
  \includegraphics[width=0.9\textwidth]{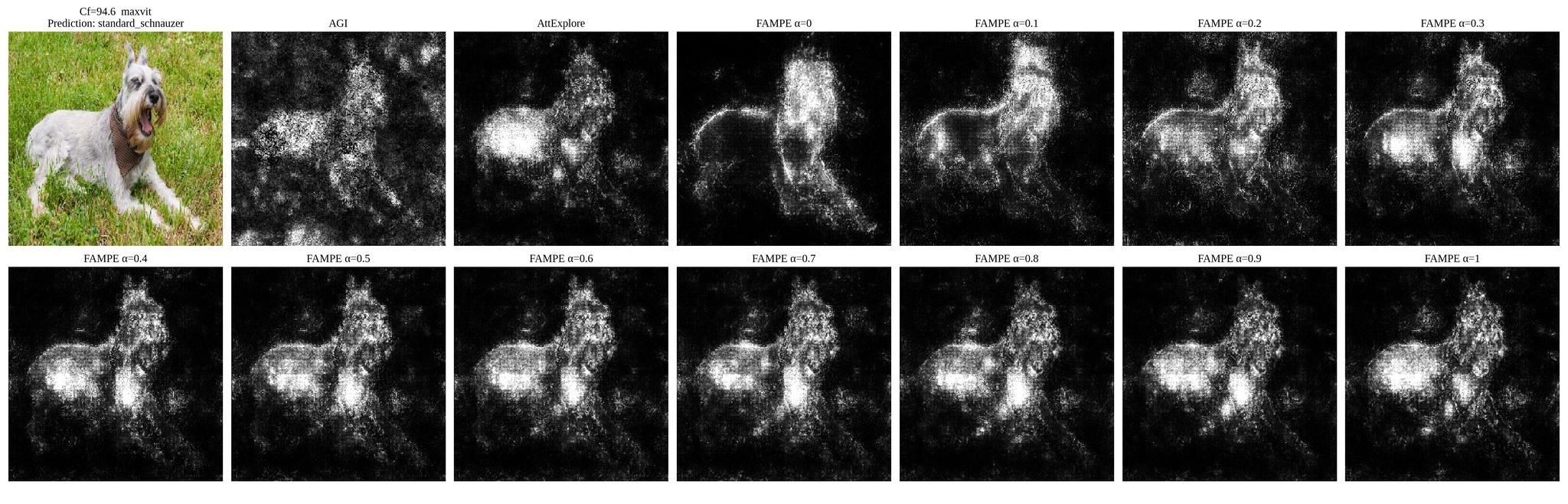}\\[2pt]
  {\small (a) Sample 9, MaxViT-T}\\[8pt]
  \includegraphics[width=0.9\textwidth]{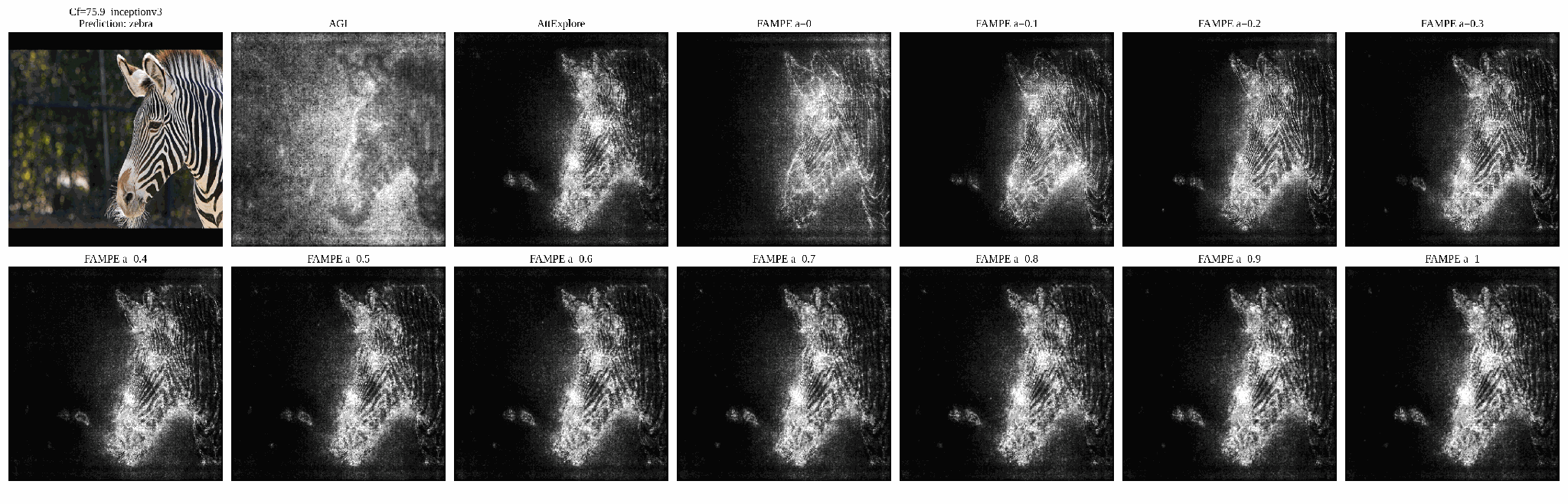}\\[2pt]
  {\small (b) Sample 10, Inception-v3}\\[8pt]
  \includegraphics[width=0.9\textwidth]{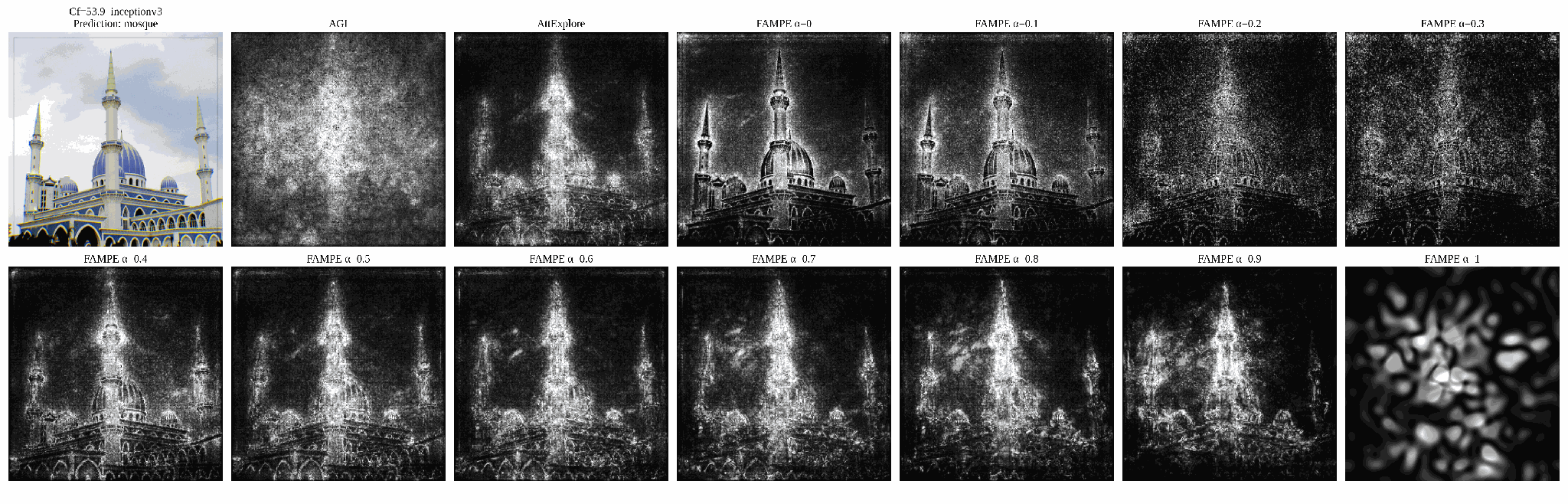}\\[2pt]
  {\small (c) Sample 11, Inception-v3}
  \caption{\newedit{Attribution maps for high-frequency images (large $c_f$), part 1. Layout as in Fig.~\ref{fig:appB_low}.}}
  \label{fig:appB_high1}
\end{figure*}

\begin{figure*}[p]
  \centering
  \includegraphics[width=0.9\textwidth]{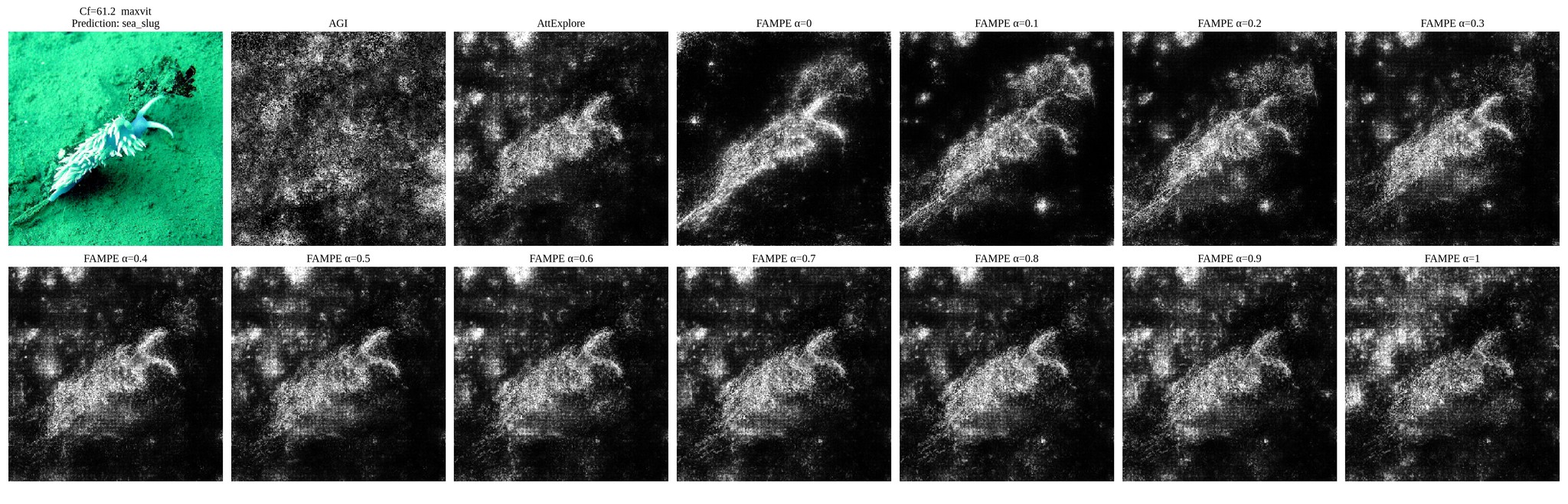}\\[2pt]
  {\small (a) Sample 12, MaxViT-T}\\[8pt]
  \includegraphics[width=0.9\textwidth]{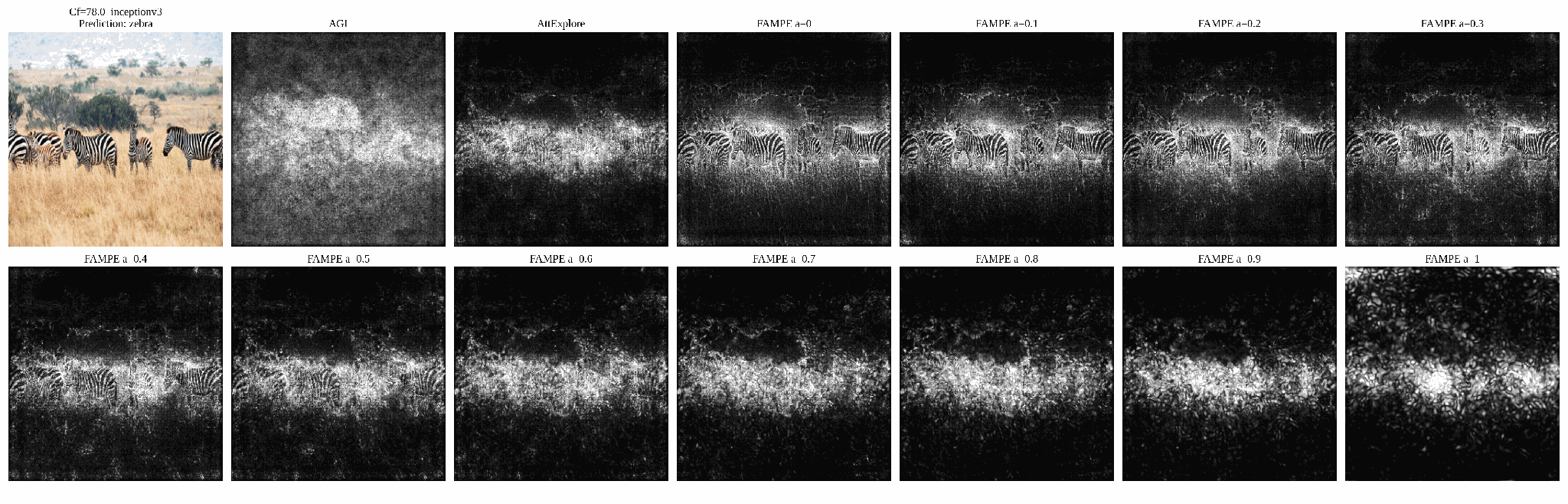}\\[2pt]
  {\small (b) Sample 13, Inception-v3}\\[8pt]
  \includegraphics[width=0.9\textwidth]{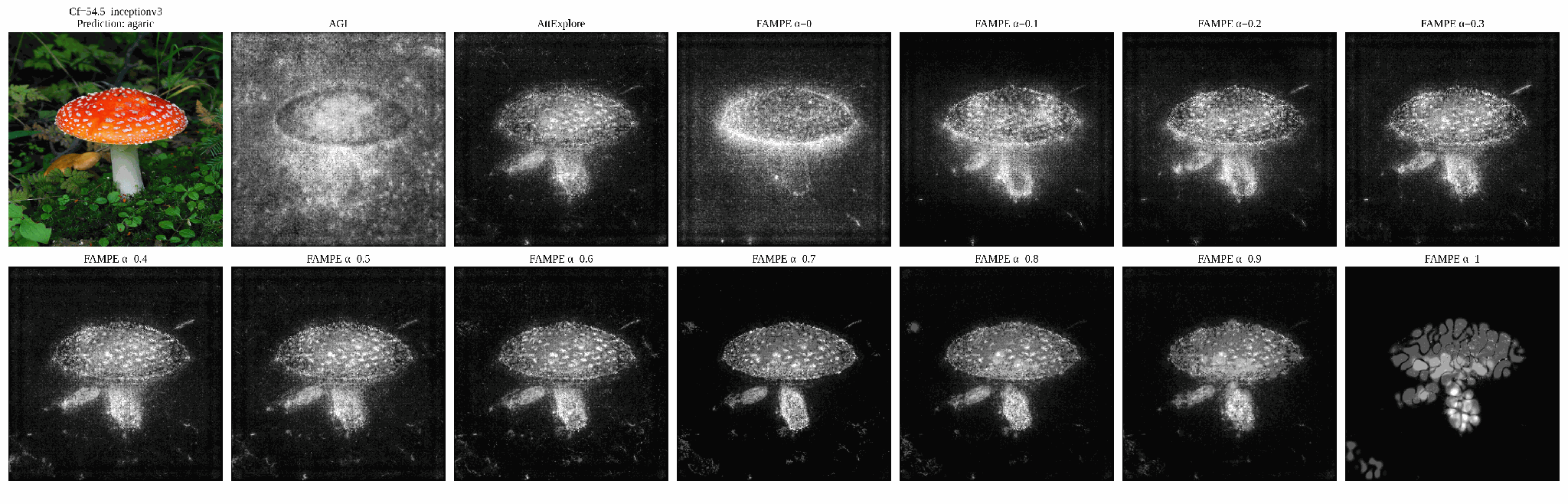}\\[2pt]
  {\small (c) Sample 14, Inception-v3}
  \caption{\newedit{Attribution maps for high-frequency images (large $c_f$), part 2. Layout as in Fig.~\ref{fig:appB_low}.}}
  \label{fig:appB_high2}
\end{figure*}

This appendix extends the qualitative comparison of Fig.~\ref{fig3}: FAMPE attribution maps across the full $\alpha$ sweep, alongside AGI and AttEXplore, for 14 additional ImageNet samples grouped by spectral cutoff $c_f$\newedit{: low (Fig.~\ref{fig:appB_low}), middle (Fig.~\ref{fig:appB_mid}), and high frequency content (Figs.~\ref{fig:appB_high1} and~\ref{fig:appB_high2}). In each panel, the input image is shown with its cutoff $c_f$, the model and the predicted class, followed by the AGI and AttEXplore maps and the FAMPE maps for $\alpha = 0, 0.1, \ldots, 1$.}

\clearpage
\bibliographystyle{elsarticle-num}
\bibliography{references}

\end{document}